\documentclass[lettersize, journal]{IEEEtran}

\usepackage[T1]{fontenc}
\usepackage{textcomp}
\usepackage{cite}
\usepackage{amsmath,amssymb,amsfonts}
\usepackage{algorithm}
\usepackage{algpseudocode}
\usepackage{graphicx}
\usepackage{subcaption} 

\begin{document}

\title{A Comparative Study on Synthetic Facial Data Generation Techniques for Face Recognition}

\author{Pedro Vidal\IEEEauthorrefmark{1}, Bernardo Biesseck\IEEEauthorrefmark{2}\IEEEauthorrefmark{1},
Luiz E. L. Coelho\IEEEauthorrefmark{3}, Roger~Granada\IEEEauthorrefmark{4}, David Menotti\IEEEauthorrefmark{1}\\
\IEEEauthorrefmark{1}{Department of Informatics, Federal University of Paran\'a, Brazil \tt\small \{pbqv20,bjgbiesseck, menotti\}@inf.ufpr.br }  \\
\IEEEauthorrefmark{2}{IFMT, Pontes e Lacerda, Brazil \tt\small \ bernardo.biesseck@ifmt.edu.br }  \\
\IEEEauthorrefmark{3}{UFMG, Brazil \tt\small \ luizcoelho@dcc.ufmg.br}  \\
\IEEEauthorrefmark{4}{Unico IDTech, Brazil \tt\small \ roger.granada@unico.io }  \\

\thanks{Preprint —This work has been submitted to IEEE for possible publication. Copyright may be transferred without notice, after which this version may no longer be accessible. Acknowledgments — We thank the \textit{Coordenação de Aperfeiçoamento de Pessoal de Nível Superior - Brasil~(CAPES)}, through the \textit{Programa de Excelência Acadêmica (PROEX)} - Finance Code 001.
This work was supported by a tripartite-contract, i.e., Unico IDTech, UFPR (Federal University of Paran\'a), and FUNPAR (Fundação da Universidade Federal do Paran\'a).  
We thank the Federal Institute of Mato Grosso (IFMT), Pontes e Lacerda, for supporting Bernardo Biesseck.
Wee also thank the National Council for Scientific and Technological Development (CNPq) (\# 315409/2023-1) for supporting Prof. David Menotti.
}
}

\maketitle


\begin{abstract}
Facial recognition has become a widely used method for authentication and identification, with applications for secure access and locating missing persons. Its success is largely attributed to deep learning, which leverages large datasets and effective loss functions to learn discriminative features. Despite these advances, facial recognition still faces challenges in explainability, demographic bias, privacy, and robustness to aging, pose variations, lighting changes, occlusions, and facial expressions. Privacy regulations have also led to the degradation of several datasets, raising legal, ethical, and privacy concerns. Synthetic facial data generation has been proposed as a promising solution. It mitigates privacy issues, enables experimentation with controlled facial attributes, alleviates demographic bias, and provides supplementary data to improve models trained on real data. This study compares the effectiveness of synthetic facial datasets generated using different techniques in facial recognition tasks. We evaluate accuracy, rank-1, rank-5, and the true positive rate at a false positive rate of 0.01\% on eight leading datasets, offering a comparative analysis not extensively explored in the literature. Results demonstrate the ability of synthetic data to capture realistic variations while emphasizing the need for further research to close the performance gap with real data. Techniques such as diffusion models, GANs, and 3D models show substantial progress; however, challenges remain.
\end{abstract}


\section{Introduction}
\label{sec:introduction}

Face recognition (FR) has rapidly become a prevalent method for authenticating and identifying users, owing to its convenience and efficiency. This technology is widely employed in various sectors, including security, where it is used to grant access to secure facilities or devices, ensuring that only authorized individuals can enter sensitive areas or use specific equipment. In law enforcement, face recognition plays a crucial role in identifying suspects, helping solve crimes, and maintaining public safety by matching faces captured in surveillance footage with criminal databases.

Additionally, it serves humanitarian purposes such as locating missing persons by comparing images with those in public records or social media. The integration of face recognition technology into everyday life is expanding at an unprecedented rate, driven by advancements in artificial intelligence (AI) and machine learning. It is now embedded in smartphones, allowing users to unlock their devices with facial recognition, and in social media platforms, where it helps tag and organize photos. 

Despite its widespread adoption and technological advancements that have propelled facial recognition into mainstream use, this technology continues to face significant challenges and controversies. One of the most pressing issues is their susceptibility to discriminatory effects and demographic bias, which can undermine their reliability and fairness. Biases often stem from unbalanced data sampling, where certain demographic groups are underrepresented, leading to skewed datasets that do not accurately reflect population diversity. The processes of data collection and labeling can also introduce biases, as they may inadvertently favor certain demographic or gender groups over others.

Moreover, FR approaches can exacerbate existing dataset biases during training because of their modeling, resulting in algorithms that perform unevenly across different demographic groups. This observation has been supported by a series of studies conducted by the US National Institute of Standards and Technology (NIST) since 2019 \cite{grother2019face,grother2021demographic}. These studies revealed significant racial and gender biases in many widely used facial recognition algorithms, highlighting that these systems often misidentify individuals from minority groups at disproportionately higher rates than those from majority groups.

In addition to bias, facial recognition technology has raised substantial privacy concerns. A major issue involves the use of web-scraped data for training models, in which images are collected without the consent of individuals. This practice can lead to unauthorized use of personal images, infringing on privacy rights, and resulting in unintended profiling. Such challenges have begun to be reflected in privacy regulations such as the EU-GDPR \cite{regulation2016regulation}, which classifies biometric data as a sensitive type of personal information subject to strict data protection rules. Consequently, some existing datasets, such as MS-Celeb1M \cite{guo2016ms} and VGGFace2 \cite{cao2018vggface2}, have been retracted because of credible privacy and ethical concerns.

Facial recognition technology has recently undergone a transformative shift with the integration of synthetic data, marking a significant evolution in the field. As traditional methods for collecting facial data have faced challenges such as privacy concerns and limited dataset availability, synthetic data has emerged as a promising solution. By generating synthetic images that replicate real-world facial features, researchers and developers can now produce vast and diverse datasets to  train facial recognition algorithms. 

Using synthetic data not only addresses ethical concerns related to the use of real individuals' data but also enables the creation of more comprehensive and representative training sets. The incorporation of synthetic data into facial recognition has attracted increasing interest from research institutions, fostering a dynamic environment for innovation and discovery. Researchers are increasingly focusing on developing facial recognition systems that are more accurate, inclusive, and privacy-aware.
Melzi \textit{et al.}~\cite{melzi2023syntheticdata}  employed their synthetic dataset GANDiffFace~\cite{melzi2023gandiffface} to reduce the gender and demographic bias of FR models pre-trained on the real faces dataset CASIA-WebFace \cite{yi2014learning}, a highly unbalanced dataset containing 41.1\% female, 58.9\% male, 84.5\% Caucasian, 11.3\% African, 2.6\% Asian, and 1.6\% Indian.

To improve the use of synthetic data and reduce the performance gap between synthetic and real datasets in facial recognition, researchers have proposed several competitions. The 1st and 2nd editions of the FRCsyn competition \cite{melzi2024frcsyn,deandres2024frcsyn} aimed to address the following questions: \textit{What are the limitations of FR technology trained solely on synthetic data?} \textit{Can synthetic data help alleviate the current limitations of the FR technology?} In the first edition, organizers proposed subtasks that invited participants to use synthetic data alone or in combination with real data to mitigate demographic bias and improve performance. In the second edition, the scope pf competition was extended to allow an unconstrained number of synthetic images, while maintaining the same objectives. Similarly, the SDFR competition \cite{shahreza2024sdfr} invited participants to submit original solutions for generating synthetic data to enhance the performance and reduce the synthetic-to-real gap.

As competitors are encouraged to submit models using either established or newly generated synthetic datasets, several important factors must be considered. For example, the dataset should exhibit sufficient intra-class variation, including changes in pose, aging, expressions, occlusions, and illumination. In addition, it must contain adequate inter-class variation to ensure that the proposed models can be generalized to unseen data. Given these challenges, generating sufficient intra-class and inter-class variation remains an active area of research. To address these challenges and provide a clear benchmark of state of the art capabilities, this study has the following objectives:

\begin{itemize}
    \item Compare the performances of face recognition models trained on various state-of-the-art synthetic datasets on verification (1:1) and identification (1:N) tasks using metrics such as accuracy, true positive rate, false positive rate, rank-1, and rank-5.
    \item A comprehensive assessment of the field is provided by contrasting various approaches and highlighting the techniques employed for this purpose.
\end{itemize}

The remainder of this paper is organized as follows. Section 2 provides an overview of the recent advances in facial recognition methods, including image preprocessing techniques, loss functions and backbone architectures. Section 3 presents the state-of-the-art synthetic data generation methods evaluated in this study. Section 4 describes our experimental methodology, details the training process and the datasets selected for evaluation. Section 5 discusses the results and compares the performance of different data generation methods. Finally, Section 6 concludes the paper, summarizes the findings, and outlines limitations identified during the experimental analysis.

\section{Advancements in 2D Face Recognition}

Facial recognition technology relies on image processing to extract features from the faces. These features are then used as inputs for pattern recognition methods to identify and match faces. These pattern recognition methods are based on machine learning, particularly deep learning networks. Deep learning has been proven to be effective in extracting information from facial images. When trained on large datasets of facial images, deep learning models can identify and extract a wide variety of facial features, such as the shape of the face, eyes, nose, mouth, and eyebrows.

Facial recognition is challenging because faces are complex and highly variable. The same face can appear differently depending on the viewing angle, lighting color and direction, and facial expressions. In addition, faces may be occluded by hair, glasses, masks, or other objects.

\subsection{Network architectures and facial recognition alignment}

2D facial recognition technology has experienced notable advancements over the years, primarily driven by the evolution of deep learning architectures. A significant breakthrough in this domain occurred with the introduction of AlexNet as proposed by Krizhevsky \textit{et al.}~\cite{krizhevsky2012imagenet}. AlexNet's success in achieving unprecedented results on the ImageNet dataset marked a turning point, leading to the widespread adoption of deep learning methods in facial recognition and paving the way for further innovation.

Following AlexNet, several key deep learning architectures have emerged, each contributing uniquely to the field. The VGGNet, developed by Simonyan \textit{et al.}~\cite{simonyan2014very}, emphasizes simplicity and depth. By employing smaller 3 $\times$ 3 convolutional filters and deeper networks, VGGNet demonstrated that increasing network depth could improve performance and influence the design of future Convolutional Neural Networks (CNN).

GoogLeNet, designed by Szegedy \textit{et al.}~\cite{szegedy2015going} and also known as Inception, introduces a novel approach with its inception module. This architecture enables more efficient computation by applying multiple filter sizes in parallel, reducing computational costs while maintaining high accuracy. The  efficiency of GoogLeNet makes it particularly suitable for large-scale applications, further advancing the capabilities of facial recognition systems.

Residual Network (ResNet), introduced in \cite{he2016deep}, addresses the problem of vanishing gradients that often affect deep networks. By incorporating residual connections, ResNet enables the training of extremely deep architectures, significantly improving the accuracy across various tasks, including facial recognition. This innovation highlights the potential of deep-learning architectures to push the boundaries of what was previously achievable.

In addition to these architectural advances, notable contributionsm specifically targeting facial recognition, have emerged. DeepFace, proposed in \cite{taigman2014deepface}, employs a nine-layer convolutional network along with a crucial facial-alignment step. This approach highlights the importance of preprocessing techniques, such as alignment, in improving recognition accuracy, demonstrating the potential of deep learning to refine facial recognition systems.

Another significant contribution came from the authors of \cite{schroff2015facenet}, with the introduction of FaceNet, which employs a convolutional network trained using triplet loss. This loss function aims to minimize the distance between an anchor sample and a positive sample (of the same class) while maximizing the distance between the anchor and a negative sample (of a different class), thereby improving the discriminative power of the learned features.

Initially, facial recognition models relied heavily on the softmax loss function combined with well-designed CNNs and large-scale training datasets. However, as the field progressed, new loss functions were developed to address challenges such as intra-class variations caused by occlusions, illumination changes, pose differences, and expression variations. A pivotal advancement occurred with the introduction of triplet loss in FaceNet~\cite{schroff2015facenet}, which provided a robust solution to these challenges by increasing the distance between the positive and negative samples using a margin factor, thus enhancing the discrimination of the learned features, as defined in~\ref{eq:triple_loss}:

\begin{equation}\label{eq:triple_loss}
\left\| f(x_a) - f(x_p) \right\|_2^2 + \alpha < \left\| f(x_a) - f(x_n) \right\|_2^2.
\end{equation}

The $x_{a}$, $x_{p}$, and $x_{n}$ represent the anchor, positive, and negative samples, respectively; $\alpha$ is the margin; and $f(\cdot)$ denotes a non linear transformation that embeds an image into a feature space.

Currently, the most prominent base loss functions are derived from Softmax, that is, SphereFace, CosFace, and ArcFace. The SphereFace loss function, proposed in \cite{liu2017sphereface}, introduced the concept of an angular margin. Unlike the traditional softmax loss, SphereFace modifies the decision boundary to include an angular margin, encouraging the network to learn features that are not only separable but also exhibit a greater angular distance between classes. However, this loss function requires a series of approximations to be computed, resulting in unstable network training. IN additional, the authors included the standard softmax loss to stabilize the training process. Empirically, softmax tends to dominate training owing to the nature of the integer-based multiplicative angular margin. This margin causes the target logit curve to become extremely steep, which makes it difficult for the model to converge effectively.

On the other hand, CosFace, introduced in \cite{wang2018cosface}, directly adds a cosine margin penalty to the target logit and achieves better performance. It also allows for a much simpler implementation and eliminates the need for joint supervision from the softmax loss. Finally, ArcFace, developed in \cite{deng2019arcface}, is one of the most widely used margin-based loss functions, as evidenced by methods submitted to competitions such as \cite{deng2021masked}. This loss function directly optimizes the geodesic distance margin of a hypersphere. This approach ensures that the learned features are not only separable, but also tightly clustered around their respective class centers, which is crucial for distinguishing between different identities. As stated in \cite{deng2019arcface}, the margin based loss functions mentioned above is defined as:

\begin{equation}
\label{eq:custom_loss2}
L = -\frac{1}{N} \sum_{i=1}^{N} \log \frac{e^{s(\cos(m_1\theta_{y_i} + m_2) - m_3)}}{e^{s(\cos(m_1\theta_{y_i} + m_2) - m_3)} + \sum\limits_{j=1, j \neq y_i}^{n} e^{s \cos \theta_j}}.
\end{equation}

The unifying principle behind these margin-based losses is to make the learned features more discriminative by enforcing a margin in the angular or cosine space, effectively pushing features from different identities further apart, while pulling features from the same identity closer together. Adjusting the parameters $m_1$ (multiplicative angular margin), $m_2$ (additive angular margin), and $m_3$ (additive cosine margin), it is possible to define different loss functions used in facial recognition:

\begin{itemize}
    \item SphereFace: Defined by \( m_1 = 1.35 \), \( m_2 = 0 \), \( m_3 = 0 \);
    \item CosFace: Defined by \( m_1 = 1 \), \( m_2 = 0 \), \( m_3 = 0.35 \);
    \item ArcFace: Defined by \( m_1 = 1 \), \( m_2 = 0.5 \), \( m_3 = 0 \).
\end{itemize}

These configurations adjust the angular margin and loss function to enhance class separability in facial recognition tasks.

\subsection{Training dataset}

In addition to the learning capacity of facial recognition models, training datasets play a crucial role in achieving a strong performance.
According to \cite{wang2021deep}, numerous face databases have been developed over the years, evolving from small to large scale, from single to diverse sources, and from controlled laboratory settings to real-world conditions. As the performance on simpler datasets, such as LFW \cite{huang2008lfw}, reached saturation, increasingly complex databases were introduced to drive progress in face recognition (FR) research. It is fair to say that the evolution of these face databases has significantly shaped the trajectory of FR research.

It is crucial that the training dataset is sufficiently large to support the development of an effective facial recognition system. In \cite{zhou2015naive}, it was proposed that the use of extensive datasets, in conjunction with deep learning techniques improves the effectiveness of face recognition. Early deep FR models were typically trained using private datasets. The DeepFace model, developed by Facebook, was trained on 4 million images from 4,000 individuals \cite{taigman2014deepface}. Similarly, Google’s FaceNet was built using a massive dataset of 200 million images representing 3 million people \cite{schroff2015facenet}. In contrast, a series of DeepID models \cite{sun2014deep,sun2015deeply,sun2014deep_2,sun2015deepid3} was trained on a smaller dataset of 200,000 images featuring 10,000 individuals. Although these models have demonstrated impressive performance at the time, researchers face challenges in replicating or effectively comparing them owing to the lack of publicly available training datasets.

The release of CASIA-WebFace \cite{yi2014learning} significantly contributed to the FR field, providing over 10,000 identities and nearly 500,000 images collected from the web using a semi-automatic approach. I became a foundational dataset for training early deep CNNs in face recognition. MS-Celeb-1M~\cite{guo2016ms} introduced a massive scale with approximately 10 million images of 100,000 celebrities, although later versions addressed noise and privacy concerns. Recently, the WebFace family of datasets has pushed the boundaries of scale and diversity. WebFace260M~\cite{zhu2021webface260m} offers increasingly larger and more refined collections, comprising over 260 million images and emphasizing cleaner annotations and demographic balance. Two curated subsets, WebFace4M and WebFace42M, were designed to meet specific research needs. WebFace4M contains 4 million images from 205,990 identities and was introduced to support the training and benchmarking of face recognition models under more controlled and noise-reduced conditions. WebFace42M, a larger subset with 42 million images across 2 million identities, was created to facilitate pretraining and scalability studies while maintaining a balance between dataset size and label quality. Glint360K~\cite{glint360k_an2022pfc}, developed by InsightFace, presents 17 million images of 360,000 identities and is widely used in conjunction with large-scale benchmarks.

Various benchmarks are commonly used to evaluate facial recognition models across variations in pose, age, illumination, and other challenging conditions. LFW~\cite{huang2008lfw} remains a widely referenced dataset containing 13,000 face images of 5,749 individuals, primarily under unconstrained settings but with limited intra-class variation. AgeDB~\cite{moschoglou2017agedb} focuses on aging and comprises 12,240 images of 440 subjects with annotated age information to test age-invariant recognition. CFP-FP~\cite{cfp_paper} targets frontal-to-profile matching with 7,000 pairs drawn from 500 subjects, highlighting performance under extreme pose changes. CPLFW~\cite{CPLFWTech} extended LFW by introducing more pronounced pose variations, while CALFW~\cite{zheng_calfw_2017} emphasizes age gaps within matched pairs. IJB-B~\cite{whitelam2017iarpa} and IJB-C~\cite{maze2018iarpa} offer more comprehensive protocols, including verification and identification tasks under video- and template-based settings with 1,845 and 3,531 subjects, respectively. TinyFace~\cite{cheng2019lowresolutionfacerecognition} introduced a low resolution face recognition benchmark with 169,403 images from 5,139 identities, designed to evaluate the performance under significant scale reduction.

\section{Synthetic Facial Data Generation Techniques} \label{sec:studied_works}

In this section, we discuss various techniques for generating synthetic faces, developed using publicly available real faces datasets. 

The methods for synthetic face generation evaluated in this study can be broadly categorized based on the underlying generative technology. These include approaches based on Generative Adversarial Networks (GANs) such as SynFace, IDnet, SFace, and DisCo; methods leveraging Diffusion Models, which have recently gained prominence, such as DCFace, Arc2Face, GANDiffFace, and IDiff-Face; techniques built on Autoencoder Architectures such as Vec2Face; and  pipelines using 3D Computer Graphics, exemplified by DigiFace.

In the following subsections, we describe the unique contributions of each. Some authors have made their synthetic datasets publicly available, which can be freely accessed and used for further research and experimentation. Consequently, the methods described herein were selected for inclusion in the experimental evaluation.

\subsection{Synthetic Datasets Publicly Available}

\subsubsection{SynFace}

\begin{figure}[!ht]
\centering
\includegraphics[width=\linewidth]{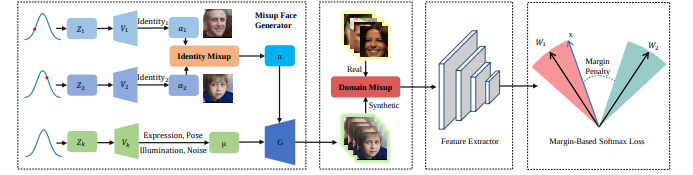}
\caption{SynFace applies identity and domain mixup to reduce the real–synthetic domain gap. Generated samples are encoded and used for margin-based loss optimization. Adapted from \cite{qiu2021synface}.}
\label{synface_pipeline}
\end{figure}

Proposed in \cite{qiu2021synface}, it explores the performance gap between models trained on synthetic and real face images, identifying limited intra-class variation and domain gaps as key contributing factors. To address these issues, the authors introduce identity mixup (IM) and domain mixup (DM) techniques. They perform sampling using a controllable face synthesis model capable of managing various factors in synthetic face generation, including pose, expression, illumination, the number of identities, and samples per identity. The overall pipeline is illustrated in Fig.~\ref{synface_pipeline}.

The generator used, named DiscoFaceGAN \cite{discoganface:2020}, synthesizes realistic face images $x$ from random noise $z$, which is composed of five independent variables $z_i \in \mathbb{R}^{N_i}$, each following a standard normal distribution. These variables represent distinct factors in face generation: identity, expression, illumination, pose, and random noise, which account for other properties, such as the background.

Let $\lambda = [\alpha, \beta, \gamma, \theta]$ denote latent factors, where $\alpha$, $\beta$, $\gamma$, and $\theta$ represent the identity, expression, illumination, and pose coefficients, respectively. Four simple Variational Autoencoders (VAEs) are then trained to map from the $z$-space to the $\lambda$-space, enabling the generator to imitate faces rendered from a 3D Morphable Model (3DMM).

The pipeline for generating a face image includes randomly sampling latent variables from a standard normal distribution, feeding them into the trained VAEs to obtain the $\alpha$, $\beta$, $\gamma$, and $\theta$ coefficients and synthesizing the corresponding face image using these coefficients. To enlarge intra-class variations, the authors propose interpolating between two different identities to create a new intermediate, formulated as:

\begin{equation}
\alpha = \phi \cdot \alpha_1 + (1 - \phi) \cdot \alpha_2,
\end{equation}

\noindent
where $\alpha_1$ and $\alpha_2$ are two randomly sampled identity coefficients from the $\lambda$-space. After sampling, the mixed face images are fed into a feature extractor, which derives the corresponding features used to compute the margin-based loss. The authors also proposed a domain mix-up technique to help mitigate the domain gap between the synthetic and real images. However, the dataset provided and collected for our evaluation did not employ this technique, and we opted not to define it.

\subsubsection{DigiFace}

\begin{figure}[!ht]
    \centering
    \begin{subfigure}[b]{0.45\textwidth}
        \centering
        \includegraphics[width=\textwidth]{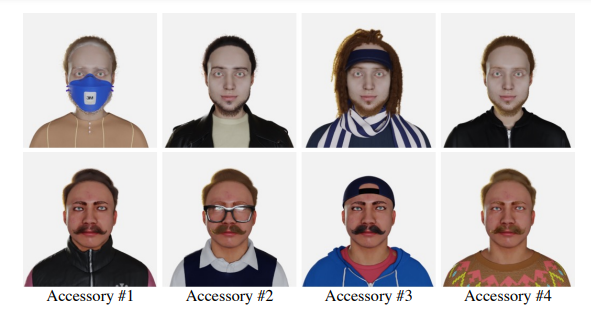}
    \end{subfigure}
    \hfill
    \begin{subfigure}[b]{0.45\textwidth}
        \centering
        \includegraphics[width=\textwidth]{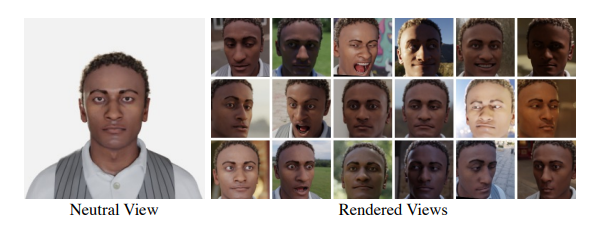}
    \end{subfigure}
    \caption{Each row presents one subject with varied accessories and hair attributes (left), and diverse pose, expression, and illumination conditions (right), improving embedding robustness. From \cite{bae2023digiface}.}
    \label{fig:data_aug_digiface}
\end{figure}

In \cite{bae2023digiface}, a state-of-the-art approach that uses a 3D model to synthesize a high-fidelity facial image dataset named DigiFace was proposed. Using a computer graphics pipeline and extensive image augmentation, the authors significantly reduced the accuracy gap between the synthetic and real data. They also fine-tuned the network with real images obtained with consent, achieving a performance comparable to models trained solely on real data. Some of the data augmentation techniques are shown in Fig.~\ref{fig:data_aug_digiface}.

\subsubsection{DCFace}

\begin{figure}[!ht]
\centering
\includegraphics[width=\linewidth]{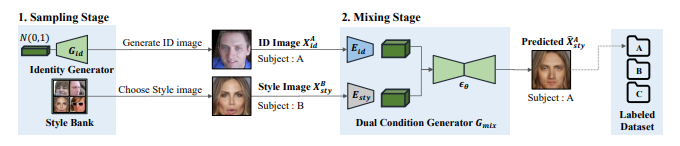}
\caption{An off-the-shelf generator provides identity and style images, which are merged by a diffusion model to produce labeled synthetic faces through repeated sampling. From~\cite{kim2023dcface}.}
\label{dcface_pipeline}
\end{figure}

In \cite{kim2023dcface}, a diffusion model pipeline consisting of two stages was proposed: (i) a sampling stage that generates a new identity image, and (ii) a mixing stage that combines the generated image with a style image from a style bank to produce a final image blending both style and identity information. Using this approach, the synthetic-to-real domain gap was significantly reduced. This method is illustrated in Fig.~\ref{dcface_pipeline}.

The two-stage data generation process employs a dual-condition generator, denoted as $G_{\text{mix}}$, which operates as a conditional Denoising Diffusion Probabilistic Model (DDPM). This generator incorporates two conditions, that is, $X_{\text{id}}$ for identity and $X_{\text{sty}}$ for style into the denoising function $\epsilon_\theta(X_t, t, E_{\text{id}}(X_{\text{id}}), E_{\text{sty}}(X_{\text{sty}}))$. These conditions are integrated through trainable feature mapping, $E_{\text{id}}$ and $E_{\text{sty}}$, and cross-attention mechanisms. The objective is $X^{A}_{\text{id}} + X^{A}_{\text{sty}} \rightarrow X^{A}_{\text{sty}}$, which uses two different images of the same subject, resulting in a mixed image that contains identity information from one image and style information from the other.

The Patch-wise Style Extractor operates on an intermediate feature of a pre-trained face recognition model, defined as $F_s(X_{\text{sty}}) = I_{\text{sty}} \in \mathbb{R}^{C \times H \times W}$. The resulting feature map is divided into a $k \times k$ grid, and each patch $I_{k_i}^{\text{sty}} \in \mathbb{R}^{C \times \frac{H}{k} \times \frac{W}{k}}$ is mapped to its mean and variance, as follows:

\begin{equation}
\hat{I}_{k_i} = \text{BN}(\text{Conv}(\text{ReLU}(\text{Dropout}(I_{k_i}^{\text{sty}})))),
\end{equation}

\begin{equation}
\mu_{k_i}^{\text{sty}} = \text{SpatialMean}(\hat{I}_{k_i}), \quad \sigma_{k_i}^{\text{sty}} = \text{SpatialStd}(\hat{I}_{k_i}),
\end{equation}

\begin{equation}
s_{k_i} = \text{LN}((W_1 \odot \mu_{k_i}^{\text{sty}} + W_2 \odot \sigma_{k_i}^{\text{sty}}) + P_{\text{emb}}),
\end{equation}

\noindent and 

\begin{equation}
E^{\text{sty}}(X_{\text{sty}}) := s = [s_1, s_2, s_{k_i}, \ldots, s_{k \times k}, s'].
\end{equation}

Here, $s'$ corresponds to $I_{k_i}^{\text{sty}}$ as a global feature, that is, when $k = 1$. $P_{\text{emb}} \in \mathbb{R}^{50 \times C}$ is a learned positional embedding \cite{gehring2017convolutional} that enables the model to extract patch styles based on their locations in $X_{\text{sty}}$. BN and LN refer to BatchNorm \cite{ioffe2015batch} and LayerNorm \cite{ba2016layer}, respectively.

To train the dual-condition generator $G_{\text{mix}}$, the original DDPM objective $L_{\text{MSE}}$, which computes the mean squared error between the true noise added to the data and the model's predicted noise, is combined with a loss that interpolates between $F(X_{\text{id}})$ and $F(X_{\text{sty}})$ across diffusion time steps, specifically:

\begin{align}
L_{\text{ID}} 
  &= -\gamma_t \, \text{CS}(F(X_{\text{id}}), F(\hat{X}_0)) \nonumber\\
  &\quad - (1 - \gamma_t) \, \text{CS}(F(X_{\text{sty}}), F(\hat{X}_0)),
\end{align}

\noindent
where $ \gamma_t \in \mathbb{R} \mid 0 \leq \gamma_t \leq 1 $. $F$ is a pre-trained face recognition model and $\text{CS}$ represents the cosine similarity and $ \hat{X}_0 $ is the new stylized image. The final loss can then be defined as:

\begin{equation}
\mathcal{L} = \mathcal{L}_{\text{MSE}} + \lambda \cdot \mathcal{L}_{ID},
\end{equation}

\noindent where $\lambda$ is a scaling parameter.

\subsubsection{IDnet}
\begin{figure}[!ht]
\centering
\includegraphics[width=\linewidth]{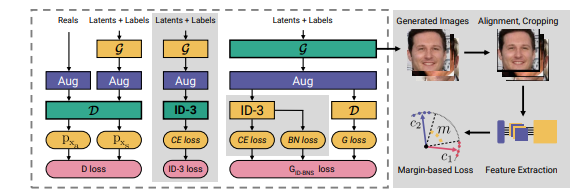}
\caption{Generator and discriminator are trained jointly with identity-driven loss and batch-norm statistics. The resulting synthetic data is processed and used to train an FR model with margin-based loss. From \cite{kolf2023identity}.}
\label{idnet_pipeline}
\end{figure}

As elaborated in \cite{kolf2023identity}, IDnet was generated using a class-conditioned StyleGAN2-ADA \cite{karras2020training}. The authors integrated the GAN \cite{goodfellow2014generative} min-max game with an identity-separable loss, named ID3, and a domain adaptation loss. This approach enables the generator to encode identity information, generate identity-separable synthetic samples, and minimize the domain gap between the synthetic and real data distributions. The pipeline is illustrated in Fig.~\ref{idnet_pipeline}.

ID-3 is trained using a margin-penalized softmax loss called CosFace~\cite{wang2018cosface}, which introduces a margin penalty to encourage training samples to cluster closely around their own class centers while remaining distant from other class centers, is defined as follows:

\begin{equation}
\label{eq_id3}
\mathcal{L}_{\text{ID-3}} = -\frac{1}{N} \sum_{i=1}^{N} \log \left(\frac{e^{s \cdot (\cos(\theta_{y_i}) - m)}}{e^{s \cdot (\cos(\theta_{y_i}) - m)} + \sum\limits_{\substack{j=1 \\ j \neq y_i}}^{C} e^{s \cdot \cos(\theta_j)}} \right),
\end{equation}

\noindent where $s$ is a scalar, $C$ is the number of classes, $N$ is the batch size, $m$ is the margin penalty, $\theta_{y_i}$ is the angle between the sample $y_i$ and its corresponding class center, and $\theta_j$ represents the angle between the sample $y_i$ and the $j$-th class center.

They also introduced a Domain Adaptation (DA) loss, intended to bridge the gap between synthetic and authentic distributions by matching the Batch Normalization Statistics (BNS) of real and synthetic data. The authors extracted the mean $\mu$ and standard deviation $\sigma$ from all the batch normalization (BN) layers of the ID\textsubscript{3} backbone. During the three-player GAN training, batch normalization statistics were computed from both the real image batch ($\mu_a$, $\sigma_a$) and synthetic image batch ($\mu_s$, $\sigma_s$). $L_{\text{BNS}}$ is used as an additional component of the default generator loss $G$. The loss terms $L_{\text{BNS}}$ and the final loss $L_{G_{\text{ID-BNS}}}$ are defined as follows:

\begin{equation}
L_{\text{BNS}}(\mu_s, \sigma_s) = \sum_{l \in \text{BNL}}\left\| \mu_s^l - \mu_a^l \right\|_2^2 + \left\| \sigma_{s}^l - \sigma_{a}^l \right\|_2^2,
\end{equation}

\noindent and

\begin{equation}
L_{G_{\text{ID-BNS}}} = L_{G_{\text{ID}}} + \lambda \cdot L_{\text{ID-3}} + \kappa \cdot L_{\text{BNS}},
\label{eq:loss_g}
\end{equation}

\noindent
where BNL refers to the batch normalization layers of ID-3, and $\lambda$ and $\kappa$ are the weighting terms for $L_{\text{ID-3}}$ and $L_{\text{BNS}}$, respectively.

\subsubsection{Arc2face}
\begin{figure}[!ht]
\centering
\includegraphics[width=\linewidth]{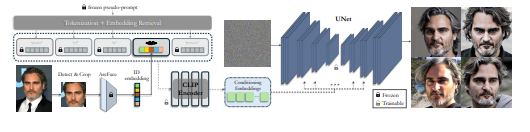}
\caption{ArcFace + CLIP embeddings guide cross-attention, with UNet and encoder trained on large-scale and high-resolution datasets without text labels. From \cite{papantoniou2024arc2face}.}
\label{arc2face_pipeline}
\end{figure}

Another approach, called Arc2face \cite{papantoniou2024arc2face}, builds on a stable diffusion model and adapts it for ID generation conditioned on identity embeddings. It focuses exclusively on ID vectors derived from ArcFace \cite{deng2019arcface}, a prominent face recognition model. This approach enables the model to generate images with strong identity consistency by fine-tuning a text encoder without requiring textual inputs. This method is illustrated in Fig.~\ref{arc2face_pipeline}.

The Arc2Face fine-tuning process serves a dual purpose: enhancing both the text encoder and diffusion model to facilitate realistic face reconstruction from abstract prompts. The text encoder is carefully fine-tuned to convert facial identity features obtained from ArcFace \cite{deng2019arcface} into the CLIP \cite{radford2021learning} embedding space used in Stable Diffusion. This involves projecting ArcFace’s 512D embeddings into a 768D space, allowing identity information to be seamlessly integrated into text representation. 
This alignment transforms the text encoder into a robust identity encoder, focusing on the facial characteristics while filtering out irrelevant contextual data, thereby ensuring identity consistency in the generated images.

Simultaneously, the diffusion model undergoes fine-tuning to align with the enhanced data conditions and identity constraints. By incorporating an expansive dataset comprising high-resolution images from sources such as a restored WebFace42M and other comprehensive datasets (e.g., FFHQ \cite{karras2019stylebased} and CelebA-HQ \cite{karras2018progressive}), the model refines its ability to understand and reconstruct diverse identity features across varying resolutions. 
The integration of identity vectors within the cross-attention layers of the UNet during training enables the diffusion model to translate noise in the latent space into coherent, identity-preserved images. This nuanced training approach ensures that the entire Arc2Face framework, from encoding to image generation, operates in concert, effectively translating abstract textual inputs into realistic facial reconstructions that maintain a subject’s unique identity traits.

\subsubsection{GANDiffFace}

\begin{figure}[!ht]
\centering
\includegraphics[width=\linewidth]{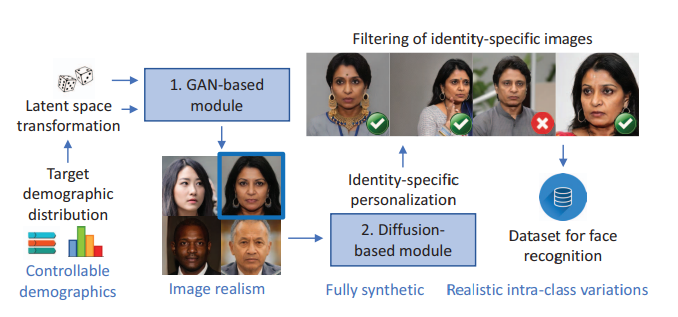}
\caption{ 
First, images with limited intra-class variations are generated using a GAN. In a second stage, such images are used to train a text-conditioned diffusion model that outputs images with realistic intra-class variations, which, once filtered, compose the final dataset. From \cite{melzi2023gandiffface}.  
}
\label{gandifface_pipeline}
\end{figure}

Another dataset, called GANDiffFace \cite{melzi2023gandiffface}, uses StyleGAN3 \cite{karras2021alias} to generate synthetic images and applies transformations in the latent space to control the generated attributes that serve as input to a diffusion model. The goal is to generate faces with desired properties, such as human face realism, controllable demographic distributions, and realistic intra-class variations. Fig.~\ref{gandifface_pipeline} illustrates the proposed pipeline.

Latent space control is based on training linear support vector machines (SVMs) \cite{cortes1995support} within the latent space to differentiate between two groups of latent vectors, each defined by a specific binary attribute, a technique initially proposed in \cite{shen2020interpreting}. The normal vector of the hyperplane created by the trained SVM indicates the direction of movement within the latent space to modify the target attributes of the facial images.

To modify a facial attribute within the latent space of StyleGAN3 \cite{karras2021alias}, adjustments can be applied to a latent vector $w$ that represents a face image, as follows:

\begin{equation}
\label{ajusts_atributes}
w_a = w + \alpha \cdot n_a,
\end{equation}

\noindent where $n_a$ is the normal vector that indicates the direction of the hyperplane separating the populations based on attribute $a$, and the factor $\alpha$ determines the extent of the modification, resulting in the altered vector $w_a$, which reflects a change in the specified attribute.

Neutralizing an attribute involves returning it to a standard or neutral state by projecting latent vector $ w $ onto the hyperplane boundary associated with that attribute. This is expressed as follows:

\begin{equation}
w_{na} = w - (w^T n_a) \cdot n_a,
\end{equation}

\noindent where $w_{na} $ is the neutralized vector, when the attribute is balanced. Using both transformation and neutralization techniques, a wide range of demographic attributes can be represented, such as race, age, and gender, which facilitate the generation of diverse identities that align with specific demographic groups.

The authors used the images generated by the GAN-based module of GANDiffFace to fine-tune Stable Diffusion, a state-of-the-art text-to-image diffusion model \cite{rombach2022high}. Training involves leveraging images produced by the GAN-based module to refine Stable Diffusion. This process uses DreamBooth to associate unique tokens, such as ``xyz,'' with each synthetic identity encoding them into the Stable Diffusion’s output space. Text prompts like ``xyz person'' refer to specific identities, utilizing the model’s prior class knowledge while preventing direct identity associations through a class-specific prior preservation loss. This regularization helps reduce overfitting and maintain image diversity. Fine-tuning involves feeding DreamBooth six images per synthetic identity and 200 regularization images into the model over 1,000 epochs, including adjustments to the text encoder, to prevent token interference across a large identity dataset and vocabulary.

Once equipped with DreamBooth, the Stable Diffusion model can generate images representing specific synthetic identities across diverse contexts using prompts that explore accessorization, advanced poses, expressions, and recontextualization. A multi-step filtering process ensures that the generated images meet quality standards: face detection is used to eliminate non-facial images; identity verification, based on ArcFace embeddings, ensures similarity scores meet a 0.3 threshold; and gender validation maintains consistency with the original textures. This comprehensive filtering process upholds the integrity and diversity of the output, thereby ensuring accuracy and consistency within the synthetic identity framework.

\subsubsection{Vec2Face}

\begin{figure}[!ht]
\centering
\includegraphics[width=\linewidth]{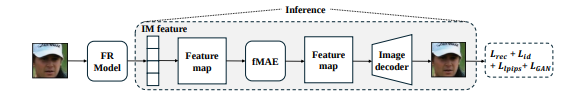}
\caption{The IM feature is computed by an FR model, expanded into a feature map, and partially masked before being decoded back to pixels. Training combines MSE, cosine similarity, perceptual, and GAN losses to preserve identity and realism. From~\cite{wu2024vec2face}.}
\label{vec2face_pipeline}
\end{figure}

In contrast, Vec2Face, developed in \cite{wu2024vec2face}, consists of a feature-masked encoder-decoder architecture. Using vectors with low similarity as inputs, different identities can be generated. In addition, by weakly perturbing the identity vector, intra-class variations are introduced. 
This method is illustrated in Fig.~\ref{vec2face_pipeline}. The authors also proposed a gradient descent method that adjusts the vector values to generate images with designated attributes.

Feature extraction is performed using a pre-trained FR model, and its features are expanded to match the input dimensions of a feature-masked auto-encoder. Similar to the Masked AutoEncoders (MAE) \cite{he2022masked}, the model is encouraged to develop improved representations by masking portions of the input data. 
Before the encoding process, a random selection of rows in the feature map was masked by $ x\% $, where $ x\% \sim \text{N}_{\text{truncated}}(\text{max} = 1, \text{min} = 0.5, \text{mean} = 0.75) $. The image Once projected, the image features are used to fill in these masked positions, thereby restoring the feature map to its original size for subsequent processing by the decoder.

Finally, the newly formed feature map is processed using an image decoder, which is responsible for generating or reconstructing the image. To enhance the quality of the resulting image, a patch-based discriminator, as described in \cite{isola2017image,yu2022scaling}, was incorporated to establish a training framework similar to a GAN.

The training objective function includes the image reconstruction loss, identity loss, perceptual loss, and GAN loss. The complete loss is defined as:

\begin{equation}
L_{\text{total}} = L_{\text{rec}} + L_{\text{id}} + L_{\text{lpips}} + L_{\text{GAN}},
\end{equation}

\noindent
where $L_{\text{rec}}$ computes the pixel-level difference between the reconstructed image and the ground-truth images. The $L_{\text{id}}$ ensures that the features extracted from the reconstructed and ground-truth images are similar. $L_{\text{lpips}}$ were used to promote the proper development of facial structures at the beginning of the training process. Additionally, a patch-based discriminator, referenced in \cite{isola2017image} and \cite{yu2022scaling}, is utilized to create the $L_{\text{GAN}}$, enhancing the sharpness of the resulting images.

During sampling, the feature vectors are controlled because larger perturbations can lead to different identities. PCA is used to sample the identity vectors that represent different identities. These identity vectors are further perturbed for image generation. Inspired by \cite{singh2023high}, the authors also introduced Explicit Attribute Control. AttrOP controls the attributes of the generated images by directly modifying the values within feature vectors through gradient descent optimization. In this manner, faces with desired attributes, such as image quality and pose, can be generated.

\subsubsection{DisCo}

\begin{figure}[!ht]
    \centering
    \begin{subfigure}[b]{0.45\textwidth}
        \centering
        \includegraphics[width=\textwidth]{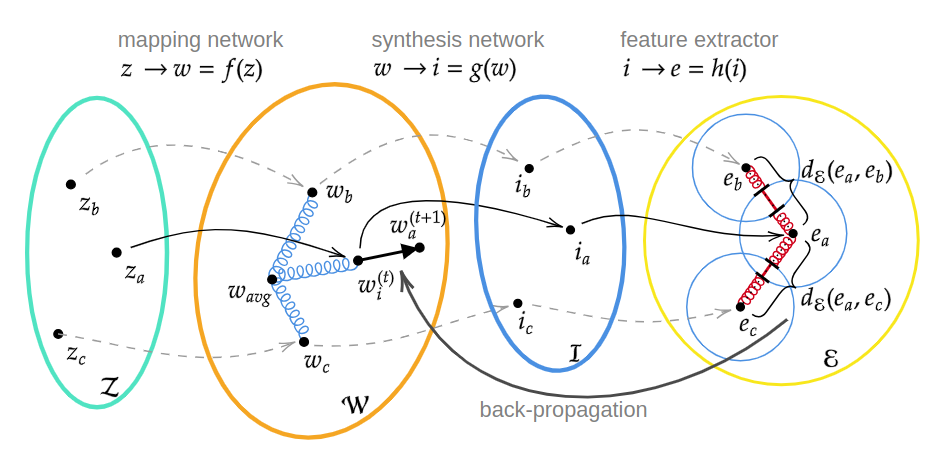}
    \end{subfigure}
    \hfill
    \begin{subfigure}[b]{0.45\textwidth}
        \centering
        \includegraphics[width=\textwidth]{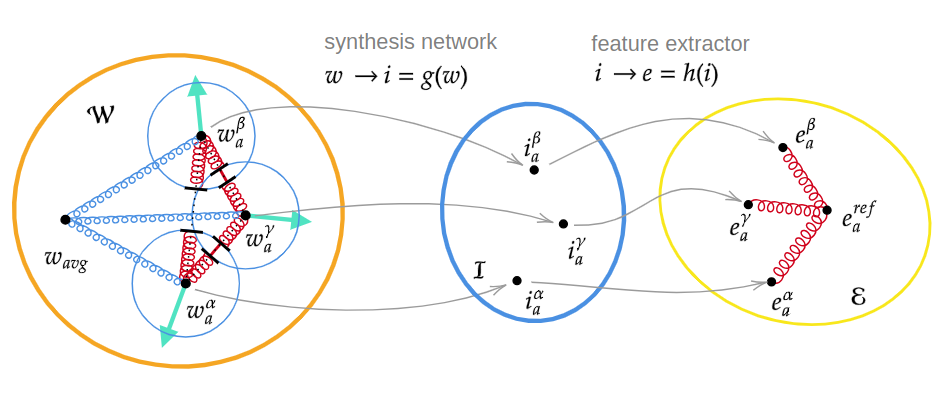}
    \end{subfigure}
    \caption{Random vectors are mapped to initial latents and iteratively refined using embedding- and latent-space losses to create inter-class variations (left). For intra-class variations (right), latents are initialized near a reference and optimized to stay close to the identity while remaining diverse. From \cite{geissbuhler2024synthetic}.}
    \label{fig:disco_pipeline}
\end{figure}

In \cite{geissbuhler2024synthetic}, a physics-inspired method was introduced to generate both intra-class and inter-class variation. Synthetic samples were produced using a Langevin-inspired iterative refinement of latent vectors, targeting an optimal distribution. Initially, embeddings were extracted from the images generated by the random latent samples. Two quadratic loss functions guide the process: one, based on granular mechanics, repels embeddings up to a threshold, and the other pulls latent vectors toward the generator’s average latent, increasing the inter-class distances while maintaining a compact latent distribution for high-quality synthesis. For intra-class variation, the Dispersion algorithm—also inspired by Langevin dynamics —acts in the latent space, adding a quadratic loss to keep embedding variability within a threshold. The initialization was further improved using a random linear combination of covariate vectors prior to refinement. The complete solution is shown in Fig.~\ref{fig:disco_pipeline}.

This is necessary because random sampling from a Gaussian distribution and using StyleGAN2 does not guarantee sufficiently diverse identities required for facial recognition model training. Thus, enhancing diversity among the generated identities is crucial. 
Generative Adversarial Networks (GANs) are used to generate synthetic identities, particularly within the StyleGAN framework, to generate diverse and useful synthetic identities for facial recognition (FR) models. The strategy involves manipulating the latent space of StyleGAN to ensure that the generated identities are sufficiently diverse and include the necessary intra-class variations.

\subsubsection{VIGFace}

\begin{figure}[!ht]
\centering
\includegraphics[width=\linewidth]{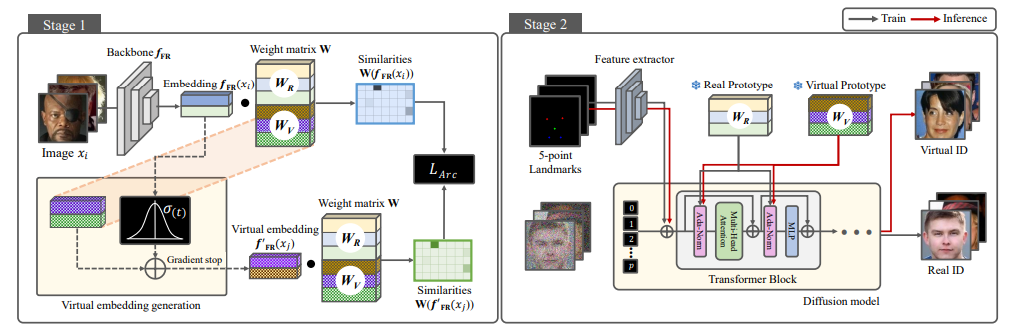}
\caption{ 
Illustration of the VIGFace pipeline, which is trained with real prototypes $WR$ and $k$ virtual ID prototypes, denoted as $WV$. The virtual embedding $f'_{FR}(x_j)$ is designed to simulate the distribution of real embeddings. Subsequently, a diffusion model is used to generate synthetic images based on the virtual prototypes. From \cite{kim2024vigface}.
}
\label{vigface_pipeline}
\end{figure}

In \cite{kim2024vigface}, VIGFace was developed, which proposed the pre-assigning of virtual identities in the feature space. The authors trained the model using both real and virtual prototypes with a prominent loss function (Arcface), thereby generating a feature space for both real and virtual identities. Subsequently, they synthesized virtual identities using a diffusion model to generate virtual entities from devised virtual prototypes. This method is illustrated in Fig.~\ref{vigface_pipeline}.

In traditional FR training methods, only prototypes for real individuals are required, which can be expressed as $W_R = [w_1^r, w_2^r, \ldots, w_n^r]$. However, in the proposed approach, the authors also incorporated additional $k$ prototypes for virtual identities, represented as $W_V = [w_1^v, w_2^v, \ldots, w_k^v]$. These images are used to create facial images of individuals who do not exist in reality. Consequently, the prototype is defined as a linear transformation matrix $W \in \mathbb{R}^{(n+k) \times D}$, where $D$ denotes the dimensions of embedding features.

To address this challenge, a virtual feature embedding, denoted as $f'_{FR}(x_j)$, was proposed to update the virtual prototypes $w_j^v$. The virtual embedding $f'_{FR}(x_j)$, associated with the virtual identity ID $j$, is generated as follows:

\begin{equation}
f'_{FR}(x_j) = w_j^v + \mathcal{N}(0, 1) \cdot \sigma,
\end{equation}

\begin{equation}
\sigma^2 = \frac{1}{b} \sum_{i=1}^{b} (f_{FR}(x_i) - w_i^r)^2,
\end{equation}

\noindent
where $\mathcal{N}(0, 1)$ represents a standard normal distribution, $\sigma$ is computed based on the variance of the embeddings, and $b$ is the mini batch size. Given that batch configuration affects the computed standard deviation, an exponential moving average (EMA) was employed to mitigate this effect. The adjusted standard deviation $\sigma$ for the current iteration $t$ was determined as follows:

\begin{equation}
\sigma = \sigma(t) \cdot \alpha + \sigma(t-1) \cdot (1 - \alpha).
\end{equation}

Here, the hyperparameter $\alpha$ is assigned a value of 0.9. The subsequent step in the process involves the use of a diffusion model to synthesize facial images. To prepare the training dataset for this model, the authors leveraged a face recognition (FR) model, which requires collecting image pairs $x_0$ along with their corresponding prototypes $w_r$. The diffusion model's inputs consist of several components: the time step $t$, FR prototype vector $w_r$, five-facial-landmark image $y$, and noisy image $x_t$. Consistent with the methodology detailed in previous research~\cite{ho2022classifier}, the model was configured to predict the velocity $v_t$ instead of the noise $\epsilon$ injected into $x_t$. The authors' model architecture was based on the DiT framework \cite{peebles2023scalable}, which draws on the principles of the visual transformer (ViT)~\cite{dosovitskiy2020image}. 
They enhanced the DiT model to incorporate images of five facial landmarks: the left eye, right eye, nose, left mouth corner, and right mouth corner \cite{zhang2016joint}. These landmark images, obtained via RetinaFace \cite{deng2019retinaface}, serve as conditioning factors to accommodate various pose variations. 

To facilitate both the generation of facial imagery and synchronization within the feature space of the FR model, the diffusion model implements a constraint that minimizes the feature distance between the original image and its corresponding prototype, as formally described as:

\begin{equation}
\label{eq:mintheta}
\min_{\theta} E_{\epsilon,t} \left\| f_{FR}\left(\hat{x}_{\theta}(x_t, t, w_r, y)\right) - w_r \right\|_2^2.
\end{equation}

Moreover, it integrates classifier-free guidance \cite{ho2022classifier} by randomly setting 10\% of the conditional embeddings $w_r$ to zero. The sampling process was as follows:

\begin{equation}
\tilde{x}_{\theta}(x_t, t, W, y) = \alpha \cdot x_{\theta}(x_t, t, W, y) + (1 - \alpha) \cdot x_{\theta}(x_t, t, y),
\end{equation}

\noindent
where $x_{\theta}(x_t, t, w_r, y)$ and $x_{\theta}(x_t, t, y)$ represent the conditional and unconditional predictions of $x_0$, respectively; and $\alpha$ denotes the guidance weight.

\subsubsection{SFace}

\begin{figure}[!ht]
\centering
\includegraphics[width=\linewidth]{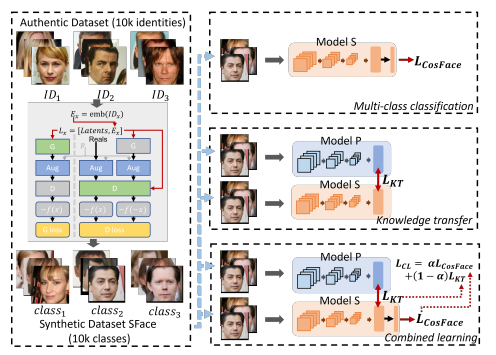}
\caption{SFace uses StyleGAN2 to generate images, conditioning the latent space with identity embeddings. Training combines multiclass classification, knowledge transfer, and joint learning. From \cite{boutros2022sface}.}
\label{sface_pipeline}
\end{figure}

Another technique proposed in \cite{boutros2022sface} is a synthetic dataset elaborated by a conditional generative adversarial network, named StyleGAN2-ADA, developed in \cite{karras2020training}. The authors used this dataset to train a facial recognition network in different settings: multiclass classification, label-free knowledge transfer, and a combination of multiclass classification and knowledge transfer. An illustrative image is shown in Fig.~\ref{sface_pipeline}.

\subsubsection{IDiff-Face}

\begin{figure}[!ht]
\centering
\includegraphics[width=\linewidth]{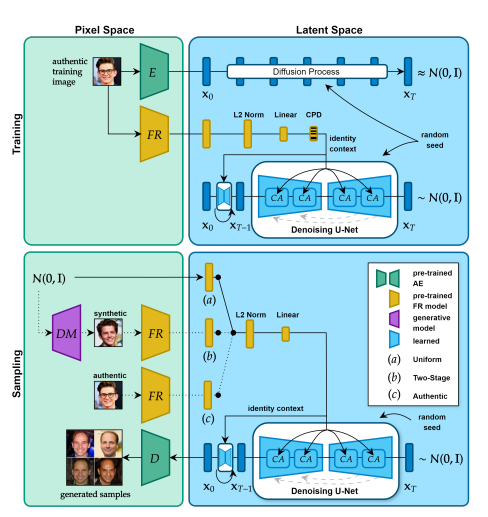}
\caption{IDiff-Face consists of a training stage, where a denoising U-Net is guided by features from a pre-trained FR model in the latent space of an autoencoder, and a sampling stage, where the trained model generates faces from real, two-stage, or synthetic identity contexts. Varying noise while fixing identity yields diverse samples. From~\cite{boutros2023idiff}.}
\label{iddiface_pipeline}
\end{figure}

Boutros et al. \cite{boutros2023idiff} proposed IDiff-Face, a diffusion model trained in the latent space of a pre-trained autoencoder and conditioned on the identity context using features extracted from a pre-trained face recognition model. Identity information is injected via cross-attention, enabling the diffusion network to focus on identity-specific traits during generations. The pipeline is illustrated in Fig.~\ref{iddiface_pipeline}.

During generation, the identity context $c$ conditions noise prediction at every time step, guiding the reverse diffusion process such that stochastic variations remain consistent with the target identity. Thus, identity conditioning influences both training and sampling, improving fidelity and maintaining coherent identity representation throughout the synthesis.

To increase intra-class diversity and avoid overfitting to the conditioning features, IDiff-Face employs Conditional Probability Dropout (CPD), which randomly drops components of the identity context during training. This prevents the model from relying heavily on a single representation and promotes variation while preserving identity consistency. As a result, IDiff-Face yields realistic identity-specific synthetic faces with enhanced intra-class variation.

\subsection{Synthetic Datasets Not Yet Publicly Available}

Some synthetic datasets are currently not publicly available, often owing to licensing restrictions or ongoing research. These datasets may contain valuable synthetic faces and were selected for inclusion in tje experimental evaluation. Future studies could explore their use as they become more accessible.

\subsubsection{$\text{ID}^3$} 

\begin{figure}[!ht]
\centering
\includegraphics[width=\linewidth]{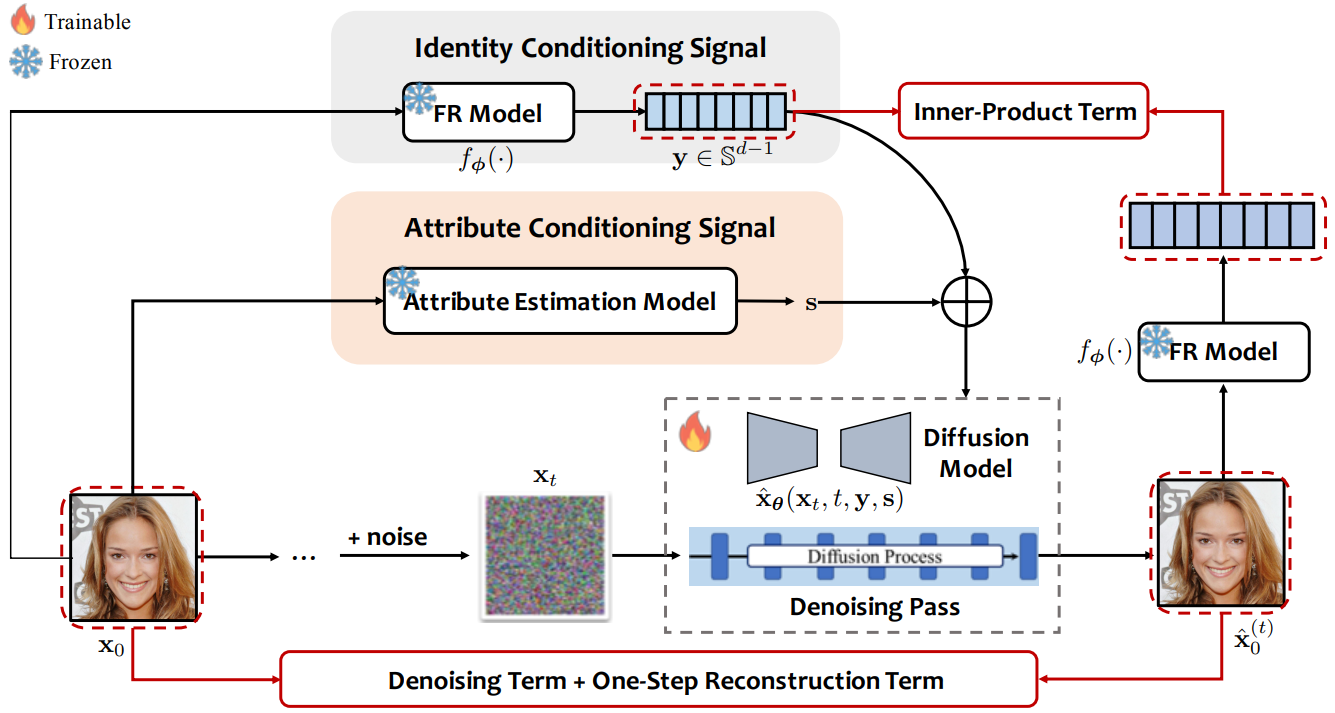}
\caption{Overview of the $\text{ID}^3$ pipeline. A face image, identity embedding, and attribute labels undergo $t$ forward diffusion steps to produce a noisy sample. A denoising network reconstructs the image conditioned on identity and attributes. Training uses a composite loss combining denoising accuracy, one-step reconstruction fidelity, identity alignment via inner product, and a constant regularization term. From \cite{id3_xu_2024}.}
\label{id3_pipeline}
\end{figure}

Proposed in \cite{id3_xu_2024}, is based on a conditional diffusion model designed to generate synthetic face datasets for training face recognition models in a privacy-preserving manner. 
The architecture of the model, presented in Fig.~\ref{id3_pipeline}, is built upon denoising diffusion probabilistic models, extending the denoising network by conditioning it on two signals: identity embeddings from a pre-trained face recognition model and facial attributes (e.g., age and pose) from pre-trained attribute predictors. This denoising network was implemented with a U-Net architecture, whereas a three-layer perceptron served as the projection module. For training, $\text{ID}^3$ primarily utilized the FFHQ dataset, and for comparative evaluations, it also incorporates the CASIA-WebFace dataset. The optimization of $\text{ID}^3$ employs a novel loss function comprising a denoising term, one-step reconstruction term, inner product term, and constant. This inner-product term specifically encourages the preservation of identity in generated images. Minimizing this proposed loss function is theoretically equivalent to maximizing the lower bound of an adjusted conditional log-likelihood over identity-preserving data, which provides an ID-preserving sampling algorithm.

\subsubsection{VariFace} 

\begin{figure}[!ht]
\centering
\includegraphics[width=\linewidth]{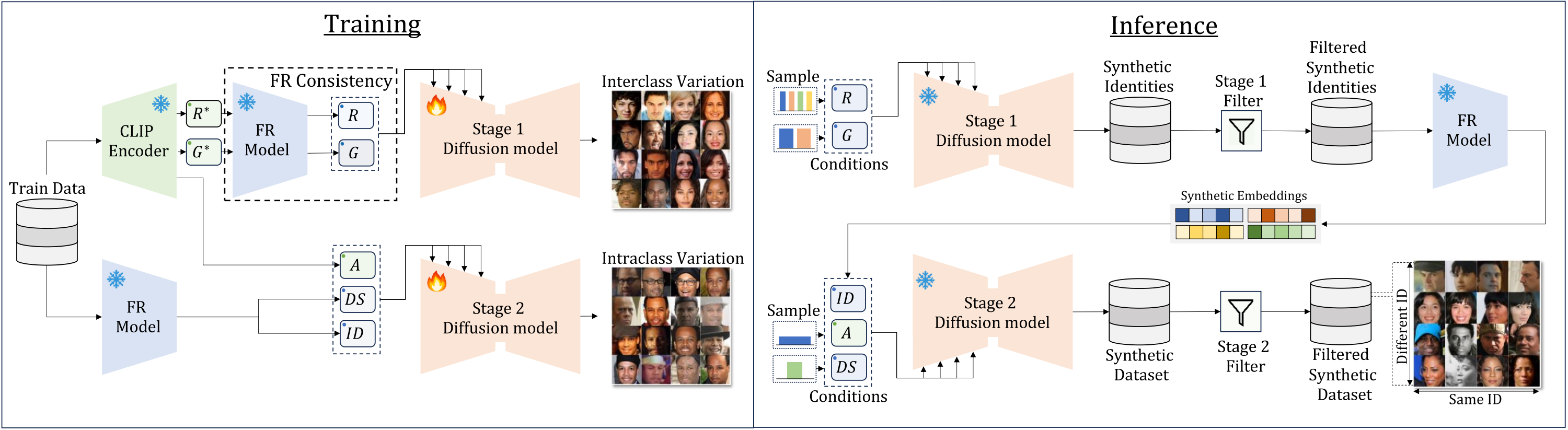}
\caption{VariFace pipeline. A pre-trained CLIP model provides initial predictions for race (R*), gender (G*), and age (A). A pre-trained FR model refines race (R) and gender (G), extracts identity (ID) embeddings, and computes divergence scores (DS). These serve as conditions to train two diffusion stages for inter- and intra-class diversity. During inference, Stage 1 generates synthetic identities, which are filtered and embedded by the FR model. Stage 2 then uses the synthetic ID embeddings, along with sampled A and DS, to synthesize faces, followed by a final quality filter. From \cite{yeung2025varifacefairdiversesynthetic}.}
\label{fig:variface_pipeline}
\end{figure}

In \cite{yeung2025varifacefairdiversesynthetic}, a two-stage diffusion-based pipeline was designed for generating fair and diverse synthetic face datasets for face recognition tasks. The architecture of the diffusion model is an adaptation of the Hourglass Diffusion Transformer (HDiT), which is modified to accommodate multiple conditioning signals, as shown in Fig.~\ref{fig:variface_pipeline}. The model was trained using the CASIA-WebFace dataset, which comprises 490,414 images across 10,575 distinct individuals. The keys to its operation are three introduced methods: Face Recognition Consistency, which refines demographic labels; Face Vendi Score Guidance, which employs the Vendi score as a guidance loss function during sampling to enhance inter-class diversity; and Divergence Score Conditioning, a metric in the face recognition embedding space that controls the balance between identity preservation and intra-class diversity. The Face Recognition models utilized within this pipeline for tasks such as label refinement and divergence score calculation, employ the ArcFace loss function.

After training, VariFace inference was performed in a two-stage process to generate synthetic faces. The first stage involves generating identity embeddings and attribute labels using two methods to enhance diversity. The first method, Face Vendi Score Guidance, is used to sample a new set of identity embeddings that are more diverse in the embedding space. This was achieved by iteratively moving the embeddings toward a higher Vendi score during the sampling process. The second method, Divergence Score Conditioning, leverages the pre-trained model to condition the sampling on a desired level of intra-class diversity, balancing it against identity preservation. The second stage of the pipeline is the diffusion-based generation of a new image. During this stage, the Hourglass Diffusion Transformer (HDiT) architecture was used, conditioned on the identity embeddings and attribute labels generated in the first stage, to produce the final synthetic face image. This two-stage approach allows VariFace to generate diverse and fair synthetic facial datasets.

\subsubsection{HyperFace} 

\begin{figure}[!ht]
\centering
\includegraphics[width=\linewidth]{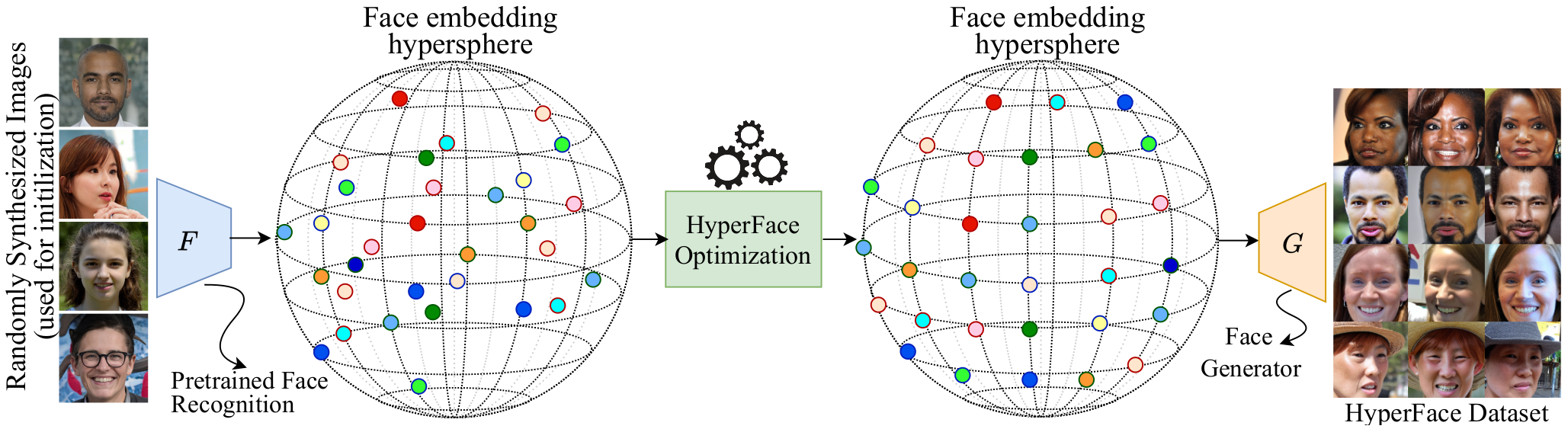}
\caption{HyperFace pipeline. Random synthetic faces are first generated, and features are extracted and normalized via a pre-trained FR model $F$ to form initial embedding points $x_{ref,i}$. HyperFace optimization then maximizes inter-class separation on the embedding hypersphere while regularizing for manifold consistency. The optimized embeddings condition a face generator $G$ to synthesize the final identities. From \cite{shahreza2025hyperface}.}
\label{fig:hyperface_pipeline}
\end{figure}

Proposed in \cite{shahreza2025hyperface} aims to maximize the inter-class variation within at synthetic dataset.
The overall architecture involves a two-stage process (see Fig.~\ref{fig:hyperface_pipeline}).
First, reference embeddings are optimized and synthetic images are generated from these embeddings. A pre-trained face recognition model, denoted as $F$ (specifically, ArcFace with an embedding dimension of 512), was used to extract identity features that cover a unit hypersphere. For image generation, a conditional face generator model, $G$, based on probabilistic diffusion models (e.g., a pre-trained generator from \cite{papantoniou2024arc2face}), synthesizes face images from the optimized embeddings and a random noise vector. The ``training data'' for the HyperFace optimization itself is initialized by generating random synthetic images using unconditional face generator models like StyleGAN or Latent Diffusion Model (LDM), from which initial reference embeddings are extracted.

The core of HyperFace is its optimization problem, which is a regularized min-max optimization derived from spherical code optimization (or Tammes problem). This optimization maximizes the minimum distance between reference embeddings on the hypersphere, ensuring a high inter-class variation. A regularization term is added to maintain the optimized embeddings on the manifold of the actual face embeddings, preventing the generation of unrealistic faces. This regularization utilizes a gallery of facial images (e.g., from StyleGAN, LDM, or real datasets suc as BUPT) to approximate the manifold of embeddings and minimize the distance of reference embeddings to this set. The optimization is solved iteratively using a gradient descent-based approach, specifically the Adam optimizer, to handle high dimensionality and a large number of identities.

\section{Datasets Evaluation}

This section describes the methodology used to perform the experiments. Section \ref{sec:image-preprocessing} presents the image preprocessing process. Section \ref{sec:training-choices} details the training choices, including the backbone used, loss function applied, data augmentation, training parameters, and training datasets derived from the methods studied in Section \ref{sec:studied_works}. 
Section \ref{sec:evaluation-datasets} describes the datasets, metrics and curves used for the  to assess the performance of the test datasets.

\subsection{Image Preprocessing} 
\label{sec:image-preprocessing}
Before training the FR models, all the faces were detected, aligned, and cropped from the original images. This was performed using the five facial landmarks automatically detected by RetinaFace \cite{deng2019retinaface}. Given a predefined set of five reference coordinate pairs for a $112 \times 112$ image, a similarity transformation matrix was computed to align the detected facial landmarks to the reference coordinates. Subsequently, the face was cropped to be centered in a $112 \times 112$ image, with the eyes aligned horizontally.

Additionally, pixel values are normalized from the range $[0, 255] \in \mathbb{Z}$ to the range $[0, 1] \in \mathbb{R}$, and standardized to have mean $\mu = 0$ and standard deviation $\sigma = 1$. Standardizing the input data typically allows the model to converge faster during training and can improve its performance.

\subsection{Training Choices}
\label{sec:training-choices}

\subsubsection{Used Backbone}
Over the past decade, CNN-based models have consistently achieved state-of-the-art performance across diverse recognition benchmarks, including face recognition, object classification, scene understanding, object detection, and other tasks, thereby solidifying their role as one of the dominant paradigms in computer vision. The basic principles of CNNs were introduced in the 1980s in \cite{lecun1989backpropagation, lecun1998gradient}, and their recent success is largely due to advances in GPU computational power, combined with the availability of large labeled datasets.

Among the numerous CNN architectures developed, Residual Networks (ResNets) \cite{he2016deep} are among the most widely adopted and influential. Their popularity stems from the introduction of residual connections, which enable effective training of very deep networks by mitigating the vanishing gradient problem. These shortcut connections allow gradients to propagate more easily through the network, facilitating convergence and improving the performance across a variety of recognition tasks.

In this study, we chose the iResNet-101 backbone, introduced in \cite{duta2021improved}, for our experiments. It is one of the top-performing backbones for deep FR \cite{deng2021masked}. The authors improved vanilla ResNet by modifying the arrangement of components and subdividing the building blocks into three stages to enhance the flow of information through the network. They also introduced a projection shortcut that reduces information loss and a convolutional block that operates on a larger number of channels, improving the performance because this block is the only component responsible for learning spatial patterns. These changes resulted in consistent improvements in the accuracy and training convergence over the baseline.

\subsubsection{Loss Function} \label{sec:loss}

Softmax loss (cross-entropy) is widely used in classification and converts logits into a probability distribution over classes, reflecting the model’s confidence. However, although effective for closed-set recognition, it provides limited feature discriminability for open-set face recognition tasks. The standard softmax loss is given by:

\begin{equation} \label{eq:softmax}
L = -\frac{1}{N} \sum_{i=1}^{N} \log \left( \frac{e^{W_{y_i}^{T} \cdot x_i + b_{y_i}}}{\sum_{j=1}^{N} e^{W_{j}^{T} \cdot x_i + b_{j}}} \right),
\end{equation}

\noindent where $x_i \in \mathbb{R}^{512}$ is the deep feature of sample $i$, $W_j$ and $b_j$ denote the class weight and bias, respectively, and $N$ and $n$ are the batch and class sizes, respectively.

To better enforce inter-class separability and intra-class compactness, angular-margin-based losses have become the state of the art in face recognition. CosFace \cite{wang2018cosface}, ArcFace \cite{deng2019arcface}, and AdaFace \cite{kim2022adaface} improved discriminability by operating in angular space. ArcFace, in particular, introduced an additive angular margin to the target logit, showing strong generalization to open-set scenarios.
Therefore, we adopted ArcFace in our experiments.

ArcFace reformulates softmax by expressing $W_j^T x_i = |W_j| \cdot |x_i| \cdot \cos(\theta_j)$ and applying $L2$ normalization,such that $|W_j| = 1$, $|x_i| = s$, and $b_j = 0$, yielding the normalized softmax as follows:

\begin{equation} \label{eq:softmax_rewrited}
L = -\frac{1}{N} \sum_{i=1}^{N} \log \left( \frac{e^{s(\cos(\theta_{y_i}))}}{e^{s(\cos(\theta_{y_i}))} + \sum_{j \neq y_i}^{C} e^{s \cos (\theta_{ji})}} \right).
\end{equation}

The final ArcFace formulation adds a margin $m$ to $\theta_{y_i}$  improving the angular discrimination, i.e.:

\begin{equation} \label{eq:arcfaceloss}
L = -\frac{1}{N} \sum_{i=1}^{N} \log \left( \frac{e^{s(\cos(\theta_{y_i} + m))}}{e^{s(\cos(\theta_{y_i} + m))} + \sum_{j \neq y_i}^{C} e^{s \cos (\theta_{ji})}} \right).
\end{equation}

\subsubsection{Data Augmentation}
\label{sec:data_augmentation}

We employed three augmentation techniques: random horizontal flip, RandAugment, and random erasing, applied sequentially using default PyTorch \texttt{transforms} settings. The random horizontal flip mirrors the image along the horizontal axis with a probability 0.5, exposing the model to appearance variations without requiring additional data collection, thereby improving generalization and reducing overfitting.

RandAugment \cite{cubuk2020randaugment} increased the data diversity by applying $N$ random transformations with magnitude $M$ selected from a predefined set (e.g., auto-contrast, equalize, rotate, solarize, color, posterize, brightness, sharpness, shear, and translation). By limiting the hyperparameters to only $N$ and $M$, RandAugment simplifies the search space while enhancing robustness to image variability.

Random erasing randomly occludes a rectangular region in the image and fill it with a constant or random value. This simulates partial occlusions and noise, thereby encouraging the model to rely on more discriminative facial regions. By introducing spatial disruptions during training, the technique improves resilience to real-world occlusions and contributes to better generalization.

For these augmentations, the default parameters from the PyTorch \texttt{transforms} module were used, and transformations were applied sequentially. Overall, the combination of these augmentations enhances the model's robustness and generalization, leading to improved accuracy.

\subsubsection{Hyperparameters}
\label{sec:hyperparameters}

All models were trained for 20 epochs with a batch size of 128 on an NVIDIA TITAN Xp GPU with 12\,GB of memory, using the Python InsightFace\footnote{https://github.com/deepinsight/insightface.} library. The Stochastic Gradient Descent (SGD) optimizer was used, with momentum set to 0.9 and weight decay set to $5 \times 10^{-4}$. The learning rate was initialized at 0.02 and decayed at each iteration according to:

\begin{equation} \label{eq:lrdecay}
\left( \frac{1.0 - \frac{l}{t}}{1.0 - \frac{l - 1}{t}} \right),
\end{equation}

\noindent where \textit{l} and \textit{t} represent the current iteration and the total number of iterations, respectively.

\subsection{Evaluation Protocol} 
\label{sec:evaluation-datasets}

Following the current face recognition literature, we evaluated the models trained on real and synthetic datasets for verification (1:1) and identification (1:N) tasks. 
The verification performance was assessed on the following test datasets: LFW~\cite{huang2008lfw}, CPLFW~\cite{CPLFWTech}, CFP-FP~\cite{cfp_paper}, CALFW~\cite{zheng_calfw_2017}, AgeDB~\cite{moschoglou2017agedb}, IJB-B~\cite{whitelam2017iarpa}, and IJB-C~\cite{maze2018iarpa}. 
For each dataset, 10-fold cross-validation was conducted and the average accuracy was reported as a percentage (\%).
For the IJB-B and IJB-C datasets, the metric used was the TPR at $\text{FPR}=0.01\%$. 
The identification performance was evaluated using rank-1 (r1) and rank-5 (r5) on the TinyFace~\cite{cheng2019lowresolutionfacerecognition} dataset. Table~\ref{tab:face_recognition_challenges} describes the datasets used for the evaluation, the number of samples, and their associated challenges.

In Section~\ref{sec:results}, we also report the Receiver Operating Characteristic (ROC) curves obtained on the LFW, CALFW, AgeDB, IJB-B, and IJB-C datasets, to observe the relationship between TPR and FPR.

Because the synthetic datasets VIGFace, VariFace, HyperFace and $\text{id}^3$ are not publicly available yet, the performances of the models trained on them were taken directly from their respective papers. 
The datasets used for training were selected based on the following  criteria: the version that achieved the best accuracy for IDiff-Face, DCFace, DigiFace, and Disco. For IDiff-Face, the Uniform version was selected, and for DCFace, the version containing 1.3M images was used. For DigiFace and Disco, the versions with the largest number of identities were chosen. Additionally, for the Arc2Face and Vec2Face datasets, versions with several images similar to those of the other datasets were selected.For the remaining methods, only one dataset version was available and used for training, and the results were obtained from the original papers and adopted in this study.

\begin{table}[!ht]
\centering
\begin{tabular}{|l|l|p{3cm}|}
\hline
\textbf{Database} & \textbf{Evaluation Protocol} & \textbf{Challenges} \\ 
\hline
LFW & 3k G and 3k I & Uncontrolled conditions, varied poses and expressions \\ 
\hline
AgeDB & 3k G and 3,000 I & Aging effects, expressions, pose variations \\ 
\hline
CFP-FP & 3k5 G and 3k5 I & Frontal-profile mismatches, pose variations \\ 
\hline
CPLFW & 3k G and 3k I & Distinct poses, cross-pose matching \\ 
\hline
CALFW & 3k G and 3k I & Cross-age facial recognition, aging effects, varied expressions, and poses \\ 
\hline
IJB-B & \text{\textasciitilde}10k G and 8M I & Variations in pose, illumination, image quality, low false positive rates \\ 
\hline
IJB-C & \text{\textasciitilde}20k G and \text{\textasciitilde}
16M I & Large-scale data, uncontrolled conditions, pose variance, low false positive rates \\ 
\hline
TinyFace & 5139 labelled identities & Low resolution FR at large scales, variations in occlusion and pose \\ 
\hline
\end{tabular}
\caption{Summary of employed test facial recognition datasets, evaluation protocols, and their challenges. G and I stand for genuines and importors.}
\label{tab:face_recognition_challenges}
\end{table}

\section{Results} \label{sec:results}

This section presents the results and a discussion of the comparative study. The advantages and disadvantages of each training dataset are also discussed.

\begin{figure}[!ht]
    \centering
    \begin{minipage}{\linewidth}
        \centering
        \includegraphics[width=\linewidth]{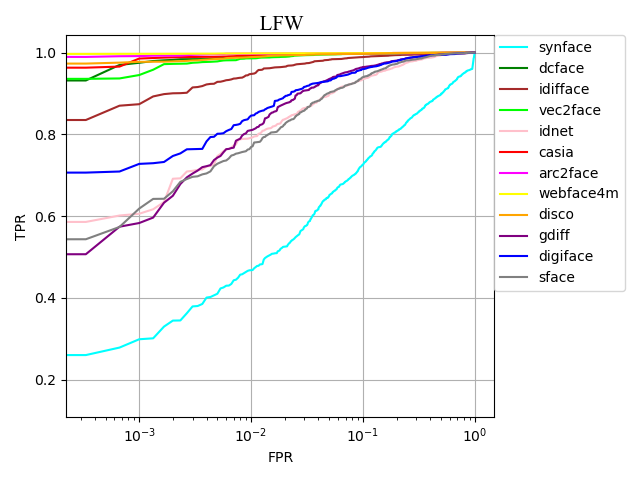}
    \end{minipage}%
    \hfill
    \begin{minipage}{\linewidth}
        \centering
        \includegraphics[width=\linewidth]{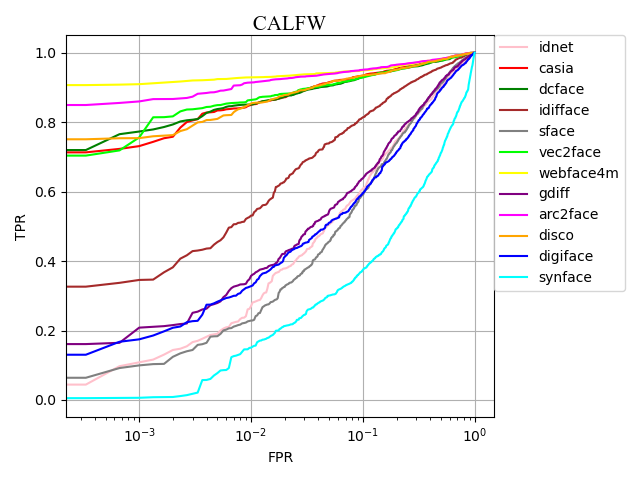}
    \end{minipage}%
    \hfill
    \begin{minipage}{\linewidth}
        \centering
        \includegraphics[width=\linewidth]{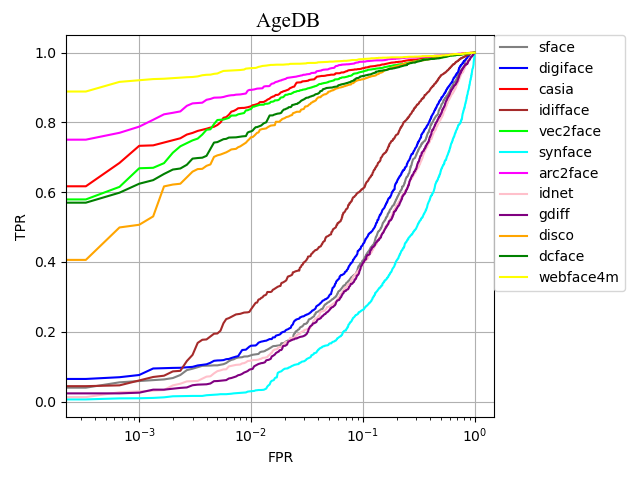}
    \end{minipage}
     \caption{ROC curves for FR models trained with real and synthetic data, evaluated on LFW, CALFW, and AgeDB (TPR: True Positive Rate, FPR: False Positive Rate).}
    \label{fig_roc_lfw_calfw_agedb}
\end{figure}

\begin{figure}[!ht]
    \centering
    \begin{minipage}{\linewidth}
        \centering
        \includegraphics[width=\linewidth]{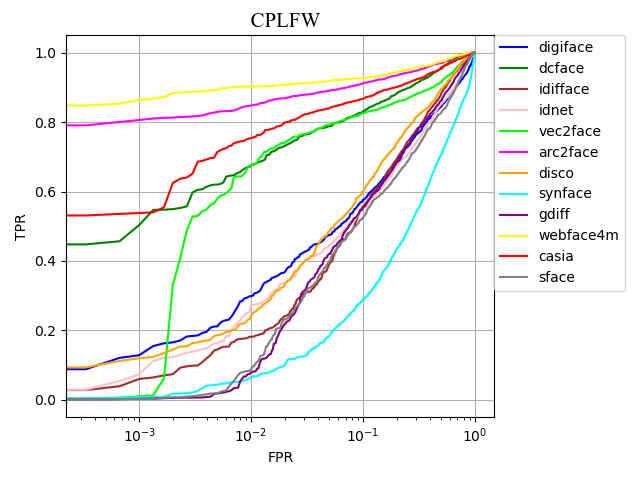}
    \end{minipage}%
    \hfill
    \begin{minipage}{\linewidth}
        \centering
        \includegraphics[width=\linewidth]{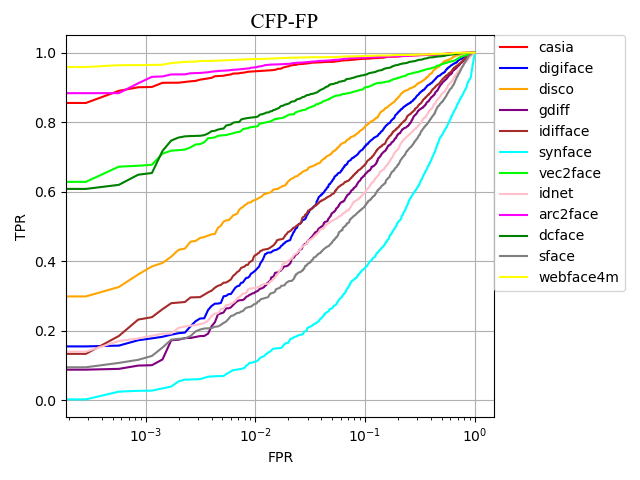}
    \end{minipage}
    \caption{ROC curves for FR models trained with real and synthetic data, evaluated on CPLFW and CFP-FP (TPR: True Positive Rate, FPR: False Positive Rate).}
    \label{fig_roc_cfpfp_cplfw}
\end{figure}

\begin{figure}[!ht]
    \centering
    \begin{minipage}{\linewidth}
        \centering
        \includegraphics[width=\linewidth]{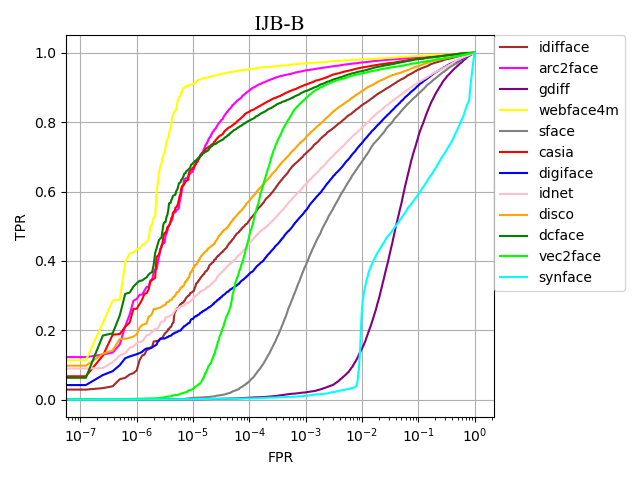}
    \end{minipage}%
    \hfill
    \begin{minipage}{\linewidth}
        \centering
        \includegraphics[width=\linewidth]{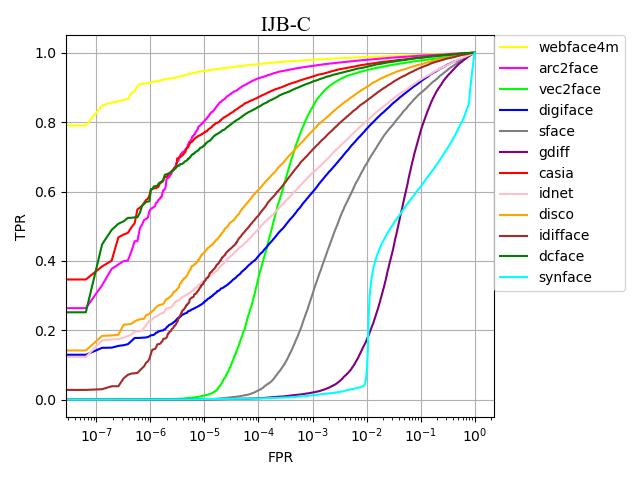}
    \end{minipage}
    \caption{ROC curves for FR models trained with real and synthetic data, evaluated on IJB-B and IJB-C (TPR: True Positive Rate, FPR: False Positive Rate).}
    \label{fig_roc_ijbb_ijbc}
\end{figure}

The results obtained for each test dataset, along with the corresponding evaluation metrics, are presented in Tables~\ref{table:results_mainstream} and~\ref{table:results_ij_tiny}. 
The ROC curves for all the verification sets are shown in Fig.~\ref{fig_roc_lfw_calfw_agedb}, Fig.~\ref{fig_roc_cfpfp_cplfw}, and Fig.~\ref{fig_roc_ijbb_ijbc}. 

For VIGFace, VariFace, HyperFace and $\text{ID}^3$, the results were obtained directly from their respective articles, as their datasets are not yet publicly available. Although the authors may have employed different training methodologies, their reported results were included for completeness. Additionally, as the articles do not report the results in datasets IJB-B, IJB-C, and TinyFace, they are omitted in Table~\ref{table:results_ij_tiny}.

\begin{table*}[!ht]
\begin{center}
    \begin{tabular}{|c|c|c|c|c|c|c|}
        \hline
        \textbf{Dataset} & \textbf{LFW} & \textbf{CPFLW} & \textbf{CFP-FP} & \textbf{CALFW} & \textbf{AgeDB} & \textbf{AVG} \\
        \hline
        WebFace4M (real)  & 99.81 & 94.68 & 98.50 & 95.91 & 97.48 & 97.28 \\
        \hline
        Arc2Face  & 99.48 & 92.53 & 97.57 & 95.21 & 95.18 & 95.99 \\
        \hline
        VariFace*  & 99.45 & 90.63 & 95.61 & 94.13 & 94.75 & 94.91 \\
        \hline
        CASIA-WebFace (real) & 99.25 & 89.65 & 97.07 & 93.33 & 94.40 & 94.74 \\
        \hline
        VIGFace* & 99.15 & 88.88 & 96.66 & 92.22 & 92.73 & 93.93 \\
        \hline
        DCFace  & 98.93 & 87.03 & 93.00 & 92.93 & 92.55 & 92.89 \\
        \hline
        Vec2Face  & 98.83 & 87.20 & 91.08 & 93.18 & 93.30 & 92.72 \\
        \hline
        HyperFace*  & 98.73 & 85.43 & 89.54 & 90.05 & 87.52 & 90.25 \\
        \hline
        $\text{ID}^3$*  & 97.68 & 82.77 & 86.84 & 90.73 & 91.00 & 89.80 \\
        \hline
        DisCo  & 99.03 & 76.53 & 84.17 & 92.98 & 91.60 & 88.86 \\
        \hline
        IDiff-Face  & 97.31 & 74.50 & 79.12 & 85.63 & 77.78 & 82.87 \\
        \hline
        DigiFace & 94.38 & 74.38 & 80.97 & 76.06 & 71.20 & 79.40 \\
        \hline
        GANDiffFace  & 94.06 & 74.38 & 78.44 & 78.30 & 68.28 & 78.69 \\
        \hline
        IDnet  & 92.58 & 73.48 & 76.08 & 77.13 & 67.96 & 77.45 \\
        \hline
        Sface  & 92.52 & 72.33 & 73.57 & 76.66 & 70.28 & 77.07 \\
        \hline
        SynFace & 81.36 & 61.78 & 65.85 & 64.10 & 60.66 & 66.75 \\
        \hline
    \end{tabular}
    \caption{Verification performance (\%) of FR models, trained on real and synthetic datasets, across the selected datasets LFW, CPLFW, CPF-FP, CALFW, and AgeDB. * indicates that the results were taken from the original article itself.}
    \label{table:results_mainstream}
 \end{center}
\end{table*}

\begin{table*}[!ht]
    \centering
    \resizebox{\textwidth}{!}{%
    \begin{tabular}{|c|c|c|c|c|c|}
        \hline
        \textbf{Dataset} & \textbf{IJB-B@0.01} & \textbf{IJB-C@0.01} & \textbf{TinyFace (r1)} & \textbf{TinyFace (r5)} & \textbf{AVG} \\
        \hline
        WebFace4M (real) & 95.18 & 96.67 & 73.81 & 76.52 & 85.55 \\
        \hline
        Arc2Face  & 89.10 & 92.66 & 65.71 & 70.41 & 79.47 \\
        \hline
        CASIA-WebFace (real)  & 82.55 & 86.66 & 57.59 & 63.65 & 72.61 \\
        \hline
        DCFace  & 79.79 & 83.56 & 53.88 & 60.13 & 69.34 \\
        \hline
        Vec2Face & 62.22 & 56.16 & 57.48 & 63.49 & 59.84 \\
        \hline
        DisCo  & 59.24 & 62.01 & 51.15 & 58.20 & 57.65 \\
        \hline
        IDiff-Face & 48.90 & 50.27 & 45.86 & 54.85 & 49.97 \\
        \hline
        DigiFace & 35.92 & 41.17 & 55.31 & 62.98 & 48.85 \\
        \hline
        IDnet  & 45.46 & 49.69 & 44.44 & 54.80 & 48.60 \\
        \hline
        Sface & 8.92 & 4.59 & 35.30 & 43.88 & 23.17 \\
        \hline
        SynFace  & 0.19 & 0.18 & 45.54 & 54.58 & 25.12 \\
        \hline
        GANDiffFace & 0.61 & 0.54 & 39.80 & 45.81 & 21.69 \\
        \hline
    \end{tabular}
    }
    \caption{Performance (\%) of FR models, trained on real and synthetic datasets, across the selected datasets IJB-B (verification, TPR@FPR=0.01\%), IJB-C (verification, TPR@FPR=0.01\%), and TinyFace (identification, rank-1 and rank-5).}
    \label{table:results_ij_tiny}
\end{table*}

Considering all datasets, the best performance was obtained by the FR model trained on WebFace4M~\cite{zhu2021webface260m}, which we considered an upper bound for the studied synthetic datasets. WebFace4M contains 4M images from 200K identities and is a subset of the larger WebFace42M dataset, which contains approximately 42 million images from 2M identities. 
This behavior was also observed in the ROC curves (see Fig.~\ref{fig_roc_lfw_calfw_agedb}, Fig.~\ref{fig_roc_cfpfp_cplfw}, and Fig.~\ref{fig_roc_ijbb_ijbc}) obtained from the verification datasets, where the yellow line consistently appeared above the others. There is one exception where the Arc2Face method slightly outperformed WebFace4M at very low FPR values for IJB-B (Fig.~\ref{fig_roc_ijbb_ijbc}), but WebFace4M surpassed it beyond that point.

The Arc2Face dataset follows, achieving an average accuracy of 95.99\% on the mainstream datasets, as shown in Table~\ref{table:results_mainstream}, and 79.47\% on IJB-B, IJB-C, and TinyFace, as shown in Table~\ref{table:results_ij_tiny}. The combination of a diffusion backbone that effectively transforms the text encoder into a face encoder, specifically tailored for projecting ArcFace embeddings into the CLIP latent space, provides a strong solution to the problem. It is scalable and is capable of generating a large number of images (i.e., the released dataset contains 21M facial images from 1M identities at a resolution of 448×448). However, it has disadvantages, such as the use of a large amount of real data (i.e., 42M images for training and 1M images for fine-tuning) for the text encoder~\cite{radford2021learning} and the diffusion backbone~\cite{rombach2022high}. Therefore, the method falls into a category that is not primarily designed to address concerns related to the use of web-scraped datasets, but rather focuses on facial attribute augmentation. This method was included in the evaluation because it  permitted submission to competition events~\cite{melzi2024frcsyn,deandres2024frcsyn,shahreza2024sdfr}.

VariFace is a 2-stage diffusion model guided by identity features (e.g., gender and age) produced by the ViT-L-14 MetaCLIP model~\cite{xu2023demystifying} and refined using iResNet-100~\cite{duta2021improved}. The FR model trained on VariFace achieved an average accuracy of 94.91\% on the mainstream datasets (Table~\ref{table:results_mainstream}). VariFace generates demographically balanced identities, which is highly relevant for reducing bias, as it is well known that celebrity datasets often exhibit imbalanced racial distributions (e.g., 84.5\% of the faces in CASIA-WebFace are Caucasian). VariFace outperformed the real dataset CASIA-WebFace except for the CFP-FP protocol (95.61\% vs. 97.07\%), likely because of the limited number of generated profile images. Considering WebFace4M as an upper bound, the synthetic-to-real gap is 2.37 percentage points (pp).

VIGFace proposed pre-assigning virtual identities in the feature space and guiding the DiT-B~\cite{peebles2023scalable} model using these virtual identities. The FR model trained on it performed close to that trained on the real dataset CASIA-WebFace, on which VIGFace was trained, with a gap of 0.81 pp.

The next evaluated dataset is DCface, a 2-stage diffusion model consisting of a sampling stage and a mixer stage. Training the FR model with DCFace yielded an average accuracy of 92.89\% on the mainstream datasets and 69.34\% on the IJB-B, IJB-C, and TinyFace datasets (Tables \ref{table:results_mainstream} and \ref{table:results_ij_tiny}). Compared to the CASIA-WebFace dataset, on which the generator was trained, the gap on mainstream datasets was 1.85 pp, while the gap on the IJB and TinyFace families is 3.27 pp. Considering the ROC curves (Fig.~\ref{fig_roc_lfw_calfw_agedb}, Fig.~\ref{fig_roc_cfpfp_cplfw}, and Fig.~\ref{fig_roc_ijbb_ijbc}), the method's curve of the mod closely followed the CASIA-WebFace curve, with the only significant difference observed in the CFP-FP protocol. They also sampled a demographically balanced dataset. This method was allowed in the 1st and 2nd editions of FRCSyn, but was not allowed in the SDFR competition \cite{shahreza2024sdfr} because it was trained using identity labels.

Vec2Face proposed a unique solution to this problem, namely, a feature-masked encoder-decoder that uses a sampled vector as input and controls the face images and their attributes. Authors trained their synthetic face generator on a subset of WebFace4M, containing 1M images. The FR model trained on Vec2Face achieved an average accuracy of 92.72\% on the mainstream datasets and 59.84\% on the IJB-B, IJB-C, and TinyFace (Tables \ref{table:results_mainstream} and \ref{table:results_ij_tiny}), respectively. It is also possible to observe that the ROC curves (Fig.~\ref{fig_roc_lfw_calfw_agedb}, Fig.~\ref{fig_roc_cfpfp_cplfw}, and Fig.~\ref{fig_roc_ijbb_ijbc}) indicate a less stable performance compared to DCFace and CASIA-WebFace.

Next is HyperFace, an optimization algorithm used to generate identity embeddings that are well distributed across a hypersphere, and later used as guidance for the Arc2Face method. However, it is possible to observe that these embeddings are less effective than those generated by PCA (i.e., the embedding sampling method for the Arc2Face approach). As an advantage, the learned embeddings are less similar to the real data compared to the PCA-generated ones, because PCA is inherently tied to the data used for analysis. With HyperFace as the training data, the FR model achieved an average accuracy of 90.25\% on mainstream datasets, with a gap of 4.49 pp compared to CASIA-WebFace (Table \ref{table:results_mainstream}).

$\text{ID}^3$ proposed a diffusion model conditioned on identity embeddings and face attribute signals (i.e., pose and age).  Enforcing identity consistency through conditioning signals and a novel loss function integrated into an existing diffusion loss, used to penalize embeddings of the generated image when distant from the target identity embedding in the cosine space, proved crucial for guaranteeing strong identity fidelity across intra-class variations and maintaining the distinctness needed for good inter-class separability. The FR model trained on $\text{ID}^3$ achieved an average accuracy of 89.80\% on the mainstream datasets with a gap of 4.94 pp compared to CASIA-WebFace (Table  \ref{table:results_mainstream}).

DisCo occurs in sequence and is based on the Brownian motion of particles to generate inter-class and intra-class variations.
Using this method, the FR model achieved an average accuracy of 88.86\% on the mainstream datasets, and an average accuracy of 57.65\% on IJB-B, IJB-C, and TinyFace, resulting in a gap to CASIA-WebFace of 5.88 pp and 14.96 pp, respectively (Tables \ref{table:results_mainstream} and \ref{table:results_ij_tiny}). Because of the use of a GAN, specifically StyleGAN2, and a low-data training regime (i.e., FFHQ), the method produced worse results than those based on diffusion models, owing to the characteristics inherent to GANs.

Next, IDiff-Face is a diffusion model conditioned the on identity context to produce identity-separable images. The FR model trained on it achieved an average accuracy of 82.87\% on the mainstream datasets and an average accuracy of 49.97\% for the IJB-B, IJB-C, and TinyFace protocols, bridging the synthetic-to-real accuracy gap to 12.1 pp and 22.64 pp when compared to CASIA-WebFace on the respective groups of datasets (Tables \ref{table:results_mainstream} and \ref{table:results_ij_tiny}). This result highlights the potential of diffusion models to generate images with variations in pose, age, expression, and illumination that, contain unique information.
As stated in \cite{kim2023dcface}, diffusion models can generate a larger number of unique identities than GANs when the sampling approach is random or guided by identity features, resulting in a dataset with greater variety. Additionally, they used a pre-trained model to extract embeddings  used by the conditional generator model based on diffusion.

DigiFace proposed a unique solution for the problem of generating synthetic faces using a computer graphics pipeline. With this approach, the FR model trained on it achieved an average accuracy of 79.40\% (Table \ref{table:results_mainstream}) on the mainstream datasets and 48.85\% (Table \ref{table:results_ij_tiny}) for the IJB-B, IJB-C, and TinyFace protocols, resulting in a synthetic-to-real accuracy gap of 15.34 pp and 23.76 pp when compared to CASIA-Webface. We observed a higher score on the CFP-FP dataset than on IDiff-Face for the FR model trained on DigiFace, suggesting that profile images are more prevalent. However, the DigiFace approach is computationally expensive and may not be feasible for research purposes. This solution is also credited without relying on large-scale real-face datasets to train certain components of its pipeline, thereby avoiding unresolved ethical issues. Nonetheless, the core challenge of generating complex synthetic data that closely mirror specific real references is analogous to the chicken-and-egg dilemma. It may be useful to reconceptualize this issue to emphasize that synthetic data inherently depend on genuine prior knowledge. Instead, focus should be placed on developing methods that make the original data difficult to reconstruct while still ensuring that the synthetic dataset adequately captures a representative view of reality \cite{geissbuhler2024synthetic}.

With GANDiffFace as training data, the FR model achieved an average accuracy of 78.69\% (Table \ref{table:results_mainstream}) on mainstream datasets and an average of 21.69 (Table \ref{table:results_ij_tiny}) on the IJB-B, IJB-C, and TinyFace protocols. The authors used StyleGAN3 to generate identity images and then fed them into a diffusion model (DreamBooth) to produce intra-class variations. The additional diffusion model enabled the dataset to exhibit a more realistic appearance than other synthetic datasets, such as IDiff-Face and DCFace, but the FR model trained on it performed worse. GANDiffFace contains demographically balanced identities; however, this comes at the cost of finetuning the diffusion model for each generated identity. The dataset also struggles with the IJB-B, IJB-C, and TinyFace protocols, showing a low TPR at a fixed FMR. 

SFace, a dataset generated using class-conditioned StyleGAN2-ADA, achieved an average accuracy of 77.07\% on mainstream datasets and an average accuracy of 23.17\% on the IJB-B, IJB-C, and TinyFace protocols (Tables \ref{table:results_mainstream} and \ref{table:results_ij_tiny}). An improved version of SFace, IdNet, achieved the same average accuracy of 77.07\% on mainstream datasets and 23.17\% on the IJB-B, IJB-C, and TinyFace protocols (Tables \ref{table:results_mainstream} and \ref{table:results_ij_tiny}). According to the authors, ``SFace suffers from relatively low identity separability, which might lead to suboptimal face verification accuracies when such synthetic data are used to train FR''~\cite{kolf2023identity}. To address this, they integrated an identity-separable loss, named ID3, into the GAN min-max game, along with a domain adaptation loss, enabling the generator to better encode identity information and generate more identity-separable images. However, this comes at the cost of requiring identity labels during training of the generative framework.

Lastly comes SynFace, which employed DiscoFaceGAN~\cite{discoganface:2020} along with identity mixup and domain mixup techniques. We observed an average accuracy of 66.75\% for the mainstream datasets and 25.12\% for the IJB-B, IJB-C, and TinyFace protocols (Tables \ref{table:results_mainstream} and \ref{table:results_ij_tiny}) for the FR model trained on SynFace. This lower performance may be due to the generator providing few unique samples. This study is also credited as one of the first to generate a synthetic dataset for training FR models.

In summary, our comparative analysis revealed several key trends in the current landscape of synthetic-face generation. First, a clear hierarchy of generative models has emerged, with diffusion-based methods (e.g., Arc2Face and DCFace) consistently outperforming their GAN-based counterparts (e.g., DisCo and SFace), demonstrating their superior ability to generate diverse and identity-preserving features.

Finally, while the synthetic-to-real performance gap is narrowing, the best methods trail the real WebFace4M by only 1-3 percentage points on some benchmarks. This gap widens considerably on more challenging, "in-the-wild" datasets like IJB-C, indicating that robustly modeling extreme variations remains the primary frontier for future research.

\section{Conclusion}

In this study, we present a comparative study of modern synthetic facial datasets generated for training face recognition models. Through evaluations across eight test benchmarks, we provide a comprehensive analysis of synthetic data generation methods that are rarely compared directly in the current literature. Our results demonstrate the progress made by techniques based on GANs, diffusion models, and 3D rendering methods aimed at capturing and generating realistic facial variations.

Approaches such as DCFace's diffusion model and SynFace's mixup methods have begun to close the performance gap between synthetic and real data, improving accuracy while addressing the ethical concerns associated with the use of real identities. These methods focus on enhancing the intra-class variation and leveraging the diversity of synthetic data to build robust facial recognition models without relying on real face data. Such technologies have the potential to reduce the need for real facial data, helping to mitigate the rights and privacy issues that arise when collecting data without explicit consent.

However, significant challenges remain, particularly regarding overfitting, computational costs, and achieving proper demographic representation in the synthesized datasets. Models such as HyperFace and Vec2Face have explored various strategies to ensure diversity and authenticity; however replicating the high performance of models trained on real data remains a challenge. Despite innovations such as VariFace and Arc2Face's identity-focused diffusion applications, synthetic datasets still face limitations in terms of realism and representation. Future research must focus on reducing the computational cost of generator models, maintaining ethical standards, and ensuring that synthetic datasets effectively capture a wide range of demographic features to serve as a viable alternative to real data in training facial recognition systems.

\section*{Acknowledgments}

No text was generated by an AI system; however, several paragraphs of the manuscript were reviewed using the ChatGPT-5 and Google Gemini AI systems.
Additionally, before the initial submission, this manuscript was checked using Paperpal Preflight software.

\bibliographystyle{IEEEtran}
\bibliography{IEEEabrv,survey}

\end{document}